%% file: main.tex
\documentclass[5p,times]{elsarticle}
\input{Sections/0_preamble}

\input{Sections/0_acronyms}

\begin{document}

\begin{frontmatter}
\title{Less Data, More Security: Advancing Cybersecurity LLMs Specialization via Resource-Efficient Domain-Adaptive Continuous Pre-training with Minimal Tokens}

\author[label1,label2,label3]{Salahuddin Salahuddin}
\author[label2]{Ahmed Hussain}
\author[label1]{Jussi Löppönen}
\author[label1]{Toni Jutila}

\affiliation[label1]{organization={SSH Communications Security},
            city={Helsinki},
            country={Finland}}

\affiliation[label2]{organization={Networked Systems Security (NSS) Group, KTH Royal Institute of Technology},
            city={Stockholm},
            country={Sweden}}
            
\affiliation[label3]{organization={Aalto University},
            city={Espoo},
            country={Finland}}

\begin{abstract}
The increasing scale of Artificial Intelligence (AI) workloads demands High-Performance Computing (HPC) infrastructure and training methodologies that are both scalable and sustainable. While Large Language Models (LLMs) demonstrate exceptional natural language capabilities, general-purpose models often lack the specialized domain knowledge necessary for effective cybersecurity analysis. In this work, we investigate Domain-Adaptive Continuous Pretraining (DAP) as a scalable, resource-efficient methodology for enhancing cybersecurity understanding in pretrained LLMs, implemented through a distributed Fully Sharded Data Parallel (FSDP) pipeline across multi-node GPU clusters. We systematically adapted three decoder-based architectures---Llama-3.1-8B, DeepSeek-R1-Distill-Qwen-14B, and Llama-3.3-70B-Instruct---using a curated 126-million-word cybersecurity corpus from standards, academic literature, and various other sources. Our approach employed constrained training parameters and distributed FSDP training to balance domain specialization with knowledge preservation. Evaluation across three cybersecurity benchmarks, namely, CTI-MCQ, CyberMetric, and SecEval, demonstrates consistent improvements post-adaptation. Notably, our Llama-3.3-70B-Ins-DAP model achieves state-of-the-art performance with accuracies of 0.718, 0.933, and 0.864, respectively, surpassing parameter-efficient baselines (accuracies $\approx$ 0.485--0.618) and specialized models including Llama-Primus-Base (trained on 2.77 billion tokens) and Foundation-Sec-8B (trained on 5 billion tokens), despite utilizing only 118.8 million tokens---representing a 23-to-42-fold reduction in training data. We demonstrate that targeted continuous pretraining, executed through scalable HPC infrastructure, enables effective cybersecurity domain adaptation with a substantially reduced computational and energy footprint, providing a foundation for specialized AI assistants in threat analysis, vulnerability assessment, and security documentation, while contributing to the broader goal of sustainable and responsible AI development.
\end{abstract}

 
\end{frontmatter}

\input{Sections/0_acronyms}
\input{Sections/1_introduction}
\input{Sections/2_preliminaries_related_works}
\input{Sections/3_methodology}
\input{Sections/4_implementaion}
\input{Sections/5_performance_evaluation}
\input{Sections/6_discussion}
\input{Sections/7_conclusion}

\bibliographystyle{elsarticle-num} 
\bibliography{main}

\end{document}

%% file: Sections/0_preamble.tex
\usepackage{amsmath,amssymb,amsfonts}
\usepackage{algorithm}
\usepackage{algpseudocode}
\usepackage{amsmath} 
\usepackage{lipsum}  
\usepackage[table,xcdraw]{xcolor}
\usepackage{graphicx}
\usepackage{enumerate}
\usepackage{tcolorbox}
\usepackage{textcomp}
\usepackage{mdframed}
\usepackage{bmpsize}
\usepackage{float}
\usepackage{pgf-pie}
\usepackage{xcolor}
\usepackage{fancyvrb}
\usepackage{multirow}
\usepackage[hang,flushmargin]{footmisc}
\usepackage{subcaption}
\usepackage{acronym}
\usepackage{enumitem}
\usepackage[acronym]{glossaries}
\usepackage{flushend}
\usepackage{comment, float, balance, tabularx, pifont}
\usepackage[normalem]{ulem}
\usepackage{tikz}
\usepackage[colorinlistoftodos,textsize=small]{todonotes}
\graphicspath{{Figures/}}
\usepackage[T1]{fontenc}
\usepackage[utf8]{inputenc}
\usetikzlibrary{chains,shapes.multipart}
\usetikzlibrary{shapes,calc}
\usetikzlibrary{automata,positioning}
\usepackage{hyperref}
\usepackage{booktabs}
\usepackage{url}                   
\newcolumntype{P}[1]{>{\centering\arraybackslash}p{#1}}

\definecolor{myLightGray}{gray}{0.9} 

\definecolor{customblue}{RGB}{0, 0, 0}
\definecolor{customred}{RGB}{0, 0, 0} 

\makeglossaries

%% file: Sections/0_acronyms.tex
\newacronym{AI}{AI}{Artificial Intelligence}
\newacronym{ML}{ML}{Machine Learning}
\newacronym{NLP}{NLP}{Natural Language Processing}
\newacronym{NLU}{NLU}{Natural Language Understanding}
\newacronym{LLM}{LLM}{Large Language Model}
\newacronym{MLM}{MLM}{Masked Language Modeling}
\newacronym{CLM}{CLM}{Causal Language Modeling}
\newacronym{PEFT}{PEFT}{Parameter Efficient Fine-Tuning}
\newacronym{KD}{KD}{Knowledge Distillation}
\newacronym{DL}{DL}{Deep Learning}
\newacronym{RNN}{RNN}{Recurrent Neural Network}
\newacronym{GRU-RNN}{GRU-RNN}{Gated Recurrent Unit Recurrent Neural Network}
\newacronym{LLaMA}{LLaMA}{Large Language Model Meta AI}
\newacronym{GPT}{GPT}{Generative Pre-trained Transformer}
\newacronym{BERT}{BERT}{Bidirectional Encoder Representations from Transformers}
\newacronym{RoBERTa}{RoBERTa}{Robustly Optimized BERT Approach}
\newacronym{LSTM}{LSTM}{Long Short-Term Memory}
\newacronym{SDN}{SDN}{Software Designed Networks}
\newacronym{IT}{IT}{Information Technology}
\newacronym{NER}{NER}{Named Entity Recognition}
\newacronym{ROC-AUC}{ROC-AUC}{Receiver Operating Characteristic Area Under the Curve}
\newacronym{DAP}{DAP}{Domain-Adaptive Continuous Pretraining}
\newacronym{RAG}{RAG}{Retrieval Augmented Generation}
\newacronym{SFT}{SFT}{Supervised Fine-Tuning}
\newacronym{RAT}{RAT}{Retrieval Aware Training}
\newacronym{CVE}{CVE}{Common Vulnerabilities and Exposures}
\newacronym{IoT}{IoT}{Internet of Things}
\newacronym{CKG}{CKG}{Cybersecurity Knowledge Graph}
\newacronym{RegEx}{RegEx}{Regular Expression}
\newacronym{EOS}{EOS}{end-of-sequence}
\newacronym{BOS}{BOS}{beginning-of-sequence}
\newacronym{GPU}{GPU}{Graphical Processing Unit}
\newacronym{CPU}{CPU}{Central Processing Unit}
\newacronym{JSONL}{JSONL}{JSON Lines}
\newacronym{LR}{LR}{Learning Rate}
\newacronym{MSL}{max\_seq\_len}{Maximum Sequence Length}
\newacronym{GAS}{grad\_accum\_step}{Gradient Accumulation Step}
\newacronym{FSDP}{FSDP}{Fully Sharded Data Parallel}
\newacronym{RL}{RL}{Reinforcement Learning}
\newacronym{FLOPS}{FLOPs}{floating point operations per second}
\newacronym{DKM}{DKM}{Domain Knowledge Mismatch}
\newacronym{GenAI}{GenAI}{Generative AI}
\newacronym{CoT}{CoT}{Chain-of-Thought}
\newacronym{CS}{CS}{Computer Science}
\newacronym{MCQ}{MCQ}{Multiple Choice Question}
\newacronym{API}{API}{Application Programming Interface}
\newacronym{CTI}{CTI}{Cyber Threat Intelligence}
\newacronym{RLHF}{RLHF}{Reinforcement Learning from Human Feedback}
\newacronym{DNN}{DNN}{Deep Neural Network}
\newacronym{LoRA}{LoRA}{Low-Rank Adaptation}
\newacronym{HPC}{HPC}{High-Performance Computing}

%% file: Sections/1_introduction.tex
\section{Introduction}
\label{sec:introduction}

\begin{figure}[!htbp]
    \centering
    \includegraphics[width=0.8\linewidth]{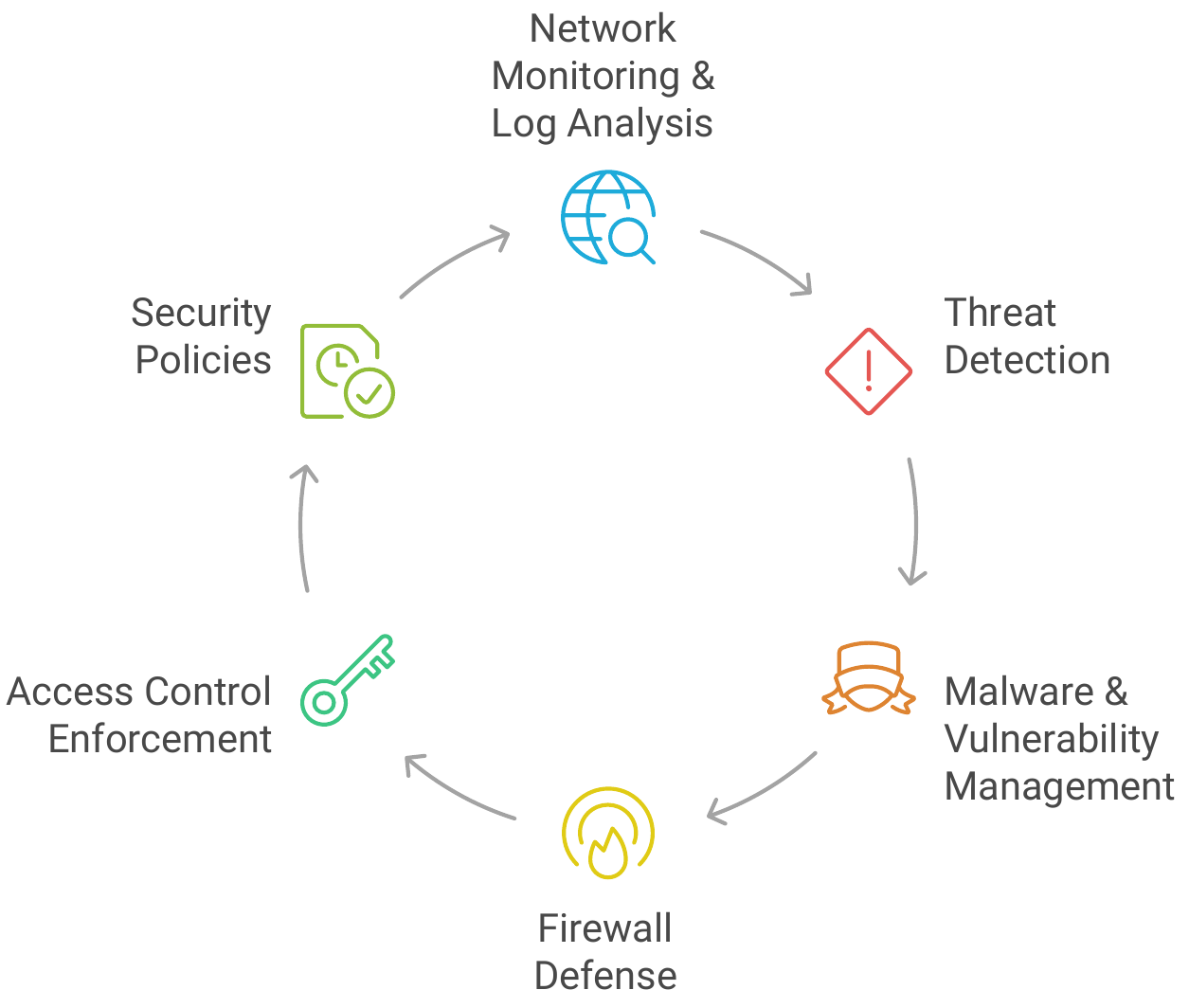}
    \caption{\color{customblue}{Traditional reactive cybersecurity framework illustrating the conventional workflow from threat detection through policy enforcement.}}
    \label{fig:traditional_approaches}
\end{figure}

The digitization of critical infrastructure, services, and data has dramatically increased the importance of cybersecurity across all sectors of society. As digital systems become increasingly complex, with numerous nodes exchanging information across networks with diverse configurations and access levels, the challenge of maintaining security grows exponentially. Traditional cybersecurity approaches, illustrated in Figure~\ref{fig:traditional_approaches}, employ reactive methodologies centered on continuous network monitoring, comprehensive log analysis, and signature-based threat detection to identify and respond to security incidents.

These conventional frameworks rely heavily on security experts who manually configure firewalls, enforce security policies, and analyze extensive datasets to prevent malware infiltration and mitigate vulnerabilities. While these established approaches provide valuable capabilities for tracking data flow and generating analytical logs, they face significant limitations when confronted with the sheer volume of information requiring real-time processing to detect sophisticated threats and vulnerabilities effectively.

The inherent reactive nature of traditional cybersecurity solutions introduces critical operational challenges. The reliance on human expertise for complex analysis tasks creates bottlenecks that are susceptible to cognitive bias, human error, and fatigue-induced oversight. Furthermore, the labor-intensive nature of manual threat assessment and policy enforcement limits scalability and response time, potentially leaving systems vulnerable during the critical window between the emergence of a threat and its detection.

Cybersecurity encompasses multiple complex domains, including wireless systems, networking infrastructure, and cloud technologies, requiring extensive expertise across various disciplines. The intricate and volatile nature of cyberspace makes it difficult to detect hidden threats without thorough analysis. Conventional methods typically identify only obvious threats, consuming a substantial amount of time and resources. 

Recent advancements in \gls{DL} architectures have revolutionized various domains, including finance, healthcare, education, and customer service. Corporations are increasingly integrating intelligent chatbots powered by \glspl{LLM}, i.e., \gls{GenAI}, to achieve greater efficiency. These agents can process vast amounts of data instantaneously, comprehend complex queries, and provide precise responses that often resolve problems without requiring human intervention. The versatility of \gls{GenAI} extends to the field of \gls{CS}, where it assists programmers and analysts with code writing and reviewing~\cite{wong2023natural}, as well as data processing and analytics~\cite{inala2024data, mitra2024generative}.

These developments in \gls{GenAI} have produced models capable of processing textual information through natural queries, employing vectorized databases for enhanced context through \gls{RAG}~\cite{lewis2020retrieval}, and providing coherent, logical responses. These characteristics make modern \glspl{LLM} particularly well-suited to enhancing security in digital systems, as they can efficiently analyze system configurations and logs, produce concise reports, and respond immediately to critical scenarios.

Despite their potential, general-purpose \glspl{LLM} lack the specialized knowledge required for effective cybersecurity analysis. We address this limitation by enhancing the cybersecurity domain knowledge of general-purpose \glspl{LLM} through \gls{DAP}, providing them with the expertise necessary to understand domain nuances while avoiding overfitting or knowledge loss. Our approach employs constrained training parameters and appropriately sized datasets to achieve this balance. Specifically, we address the following: 
\begin{tcolorbox}
{\small
\textbf{Research Question:} \textit{Can efficient domain adaptation enable cybersecurity specialization of pretrained \glspl{LLM} using substantially smaller datasets than traditional pretraining approaches?}
}
\end{tcolorbox}

The pursuit of effective \gls{LLM} specialization aligns directly with the imperatives of scalable, sustainable \gls{AI} development. Training and adapting large-scale language models demand significant computational resources, often relying on \gls{HPC} infrastructure to remain feasible. At the same time, the rising energy costs of \gls{AI} workloads have elevated resource efficiency from a practical convenience to an ethical and environmental imperative. 

Our investigation addresses both dimensions simultaneously: by demonstrating that competitive cybersecurity specialization is achievable with substantially smaller datasets than current approaches assume, we contribute a methodology that reduces computational overhead, lowers energy consumption, and broadens access to domain-specific \gls{AI} capabilities for resource-constrained academic and industrial organizations. The distributed \gls{FSDP} training pipeline we employ is designed for scalability across multi-node \gls{GPU} clusters, and the data efficiency findings we report have direct implications for sustainable \gls{AI} practices at scale.

To investigate this systematically, we undertake four principal objectives: (i) preparing a comprehensive cybersecurity corpus, (ii) selecting appropriate open-access \glspl{LLM} for adaptation, (iii) implementing an efficient domain-adaptation pipeline, and (iv) rigorously assessing the resulting models through established benchmark datasets.

\textbf{Contribution.} This paper addresses the gap between general-purpose \glspl{LLM} and domain-specific cybersecurity requirements through a novel systematic domain adaptation methodology. Our investigation demonstrates that specialized cybersecurity knowledge can be effectively instilled in pretrained models using carefully curated, resource-efficient training approaches. Our primary contributions are summarized as follows:

\begin{enumerate}[leftmargin=*]
    \item We develop a systematic methodology for cybersecurity domain adaptation by instilling specialized knowledge through constrained training parameters, conservative learning rates, and limited training epochs (2-3).
    
    \item We curate a specialized 126-million-word cybersecurity dataset from authoritative standards (e.g., ISO, NIST), academic literature, and technical documentation, providing comprehensive domain coverage while maintaining computational feasibility for domain adaptation.
    
    \item We demonstrate competitive performance with substantially smaller datasets (118.8 million tokens) than existing specialized models (\textcolor{customblue}{up to 5 billion} tokens), challenging prevailing assumptions about data requirements for effective \gls{LLM} domain specialization.
    
    \item We systematically evaluate domain adaptation effectiveness across three decoder-based architectures, providing empirical insights into scale-dependent learning dynamics and optimal resource allocation strategies.
    
    \item {\color{customred}We provide empirical evidence demonstrating that \gls{DAP} significantly outperforms parameter-efficient baselines (\gls{PEFT}/\gls{LoRA}) in cybersecurity tasks.}
\end{enumerate}

Our domain-adapted models achieve higher performance across established benchmarks (CTI-MCQ, CyberMetric, SecEval), with the \textit{Llama-3.3-70B-Ins-DAP} model obtaining accuracies of 0.718, 0.933, and 0.864, respectively, surpassing specialized cybersecurity models including \textit{Llama-Primus-Base}~\cite{yu2025primus} \textcolor{customblue}{and \textit{Foundation-Sec-8B}~\cite{kassianik2025llama}}. Additionally, our approach introduces key technical advances, where we utilize \gls{FSDP} technique for computational scalability, \textcolor{customblue}{conservative hyperparameter selection to mitigate catastrophic forgetting while enabling effective knowledge acquisition}, and a comprehensive evaluation framework spanning multiple cybersecurity subdomains. These contributions establish foundational capabilities for developing specialized cybersecurity \gls{AI} assistants capable of threat analysis, vulnerability assessment, and security documentation generation.

\textbf{Paper Organization.}  Section~\ref{sec:preliminaries_related_works} provides the theoretical foundation covering transformer architectures, domain adaptation methodologies, and \gls{GenAI} applications in cybersecurity. Sections~\ref{sec:methodology} and~\ref{sec:implementation} present our systematic domain adaptation framework and distributed training implementation. Section~\ref{sec:perf_eval} evaluates model performance through comprehensive benchmarking against baseline and specialized cybersecurity models. Section~\ref{sec:discussion} synthesizes empirical findings within the broader domain-specific \gls{LLM} research landscape. Section~\ref{sec:conclusion} consolidates contributions and identifies future research directions in efficient domain adaptation methodologies.

%% file: Sections/2_preliminaries_related_works.tex
\section{Preliminaries and Related Work}
\label{sec:preliminaries_related_works}

This section establishes the conceptual foundations necessary for understanding the domain adaptation of \glspl{LLM} for cybersecurity applications. We begin by examining the evolution of language understanding technologies, proceed to analyze transformer architectures that power modern language models, explore domain adaptation methodologies, and review relevant research in applying these technologies to cybersecurity challenges. Following this, we introduce an evaluation framework comprising five dimensions for comparing \glspl{LLM} domain adaptation in cybersecurity. Finally, we review existing research in cybersecurity-focused language model adaptation, organizing approaches by architectural paradigm and identifying research gaps that motivate our investigation into data-efficient domain specialization.

\subsection{Language Understanding}
\label{sec:pre_lang_under}

\gls{NLP} represents a core domain of \gls{AI} focused on processing natural text for decision-making applications. Traditional statistical techniques such as N-Gram and TF-IDF have been employed for tasks including text classification, information retrieval, and language modeling~\cite{yun2005improved, ramos2003using, brown1992class}. Despite their utility for applications such as spam filtering, these classical approaches exhibit significant limitations: they struggle with high-dimensional data, demonstrate vulnerability to data sparsity, and lack a meaningful contextual understanding of semantic relationships. The emergence of \gls{DL} architectures enabled a fundamental change in \gls{NLP}. \glspl{DNN} such as \glspl{RNN}~\cite{elman1990finding} and \gls{LSTM}~\cite{hochreiter1997long} transcended the constraints of statistical methods by efficiently processing large text corpora and extracting semantic meaning. These architectures have demonstrated exceptional performance across diverse applications, including speech recognition~\cite{graves2013speech}, sequence-to-sequence modeling for machine translation~\cite{sutskever2014sequence}, and \gls{NER}~\cite{lample2016neural}.

\subsection{Transformer Architectures}
\label{sec:pre_trans_arch}

The transformer architecture introduced by Vaswani et al.~\cite{vaswani2017attention} fundamentally revolutionized language processing. Unlike previous sequential processing methods, transformers use self-attention mechanisms within an encoder-decoder framework, enabling parallel text processing and generating contextually rich representations. Transformer models generate output by predicting sequential tokens until reaching a designated end token or the model's context limit. For each prediction, the model produces a probability distribution across its vocabulary, selecting the most likely token via predefined algorithms. While context windows constrain token generation capacity, various methodologies exist to expand effective context, including input truncation and positional embedding adjustments~\cite{ding2024longrope}.

\textbf{Encoder-Decoder Framework.} The transformer architecture's encoder-decoder framework enables parallel processing through multi-head attention mechanisms. The encoder transforms input text into high-dimensional, context-rich internal representations (hidden states), which the decoder then processes to generate appropriate output text for tasks such as language translation~\cite{wang2019learning}. During training, input sequences pass through an embedding layer that converts text into dense, uniform-length vectors, capturing semantic and syntactic characteristics. These embeddings traverse attention mechanisms and feed-forward networks to produce context-enriched encodings, which the decoder's cross-attention mechanism subsequently utilizes. The decoder's masked self-attention ensures that each token attends only to itself and previously generated tokens, enabling coherent response generation.

\subsection{Large Language Models}
\label{sec:pre_llms}
\glspl{LLM} typically employ transformer architectures that are pretrained on extensive text corpora collected from diverse sources. While maintaining architectural similarities to standard transformers, these models incorporate numerous attention layers that enable the extraction of complex contextual and syntactic information. Contemporary \glspl{LLM} generally utilize either encoder-only or decoder-only architectures, although some implementations, such as T5~\cite{raffel2020exploring} and BART~\cite{lewis2019bart}, leverage both components. The T5 approach, for instance, conceptualizes all natural language tasks as text-to-text transformations, enabling fine-tuning for specific applications, such as question answering or summarization, following general pretraining.

\textbf{Encoder-only Models.} Encoder-only architectures stack multiple transformer encoder components, as exemplified by \gls{BERT}~\cite{devlin2018bert} and \gls{RoBERTa}~\cite{liu2019roberta}. These models transform raw text into semantically rich encodings that capture the relationships between tokens. Such architectures excel at tasks requiring deep textual understanding for contextual prediction, including \gls{NER} and sentiment analysis~\cite{tan2022roberta}. Encoder-based models typically train through \gls{MLM}, wherein certain tokens are randomly masked before embedding, with the model learning to infer these masked elements. This approach enables the model to develop robust semantic representations that can subsequently support classification tasks~\cite{lan2019albert}.

\textbf{Decoder-only Models.} Decoder-only models specialize in generative tasks, including text generation and language modeling. Unlike encoder-only architectures that primarily encode contextual information, these models learn to produce coherent, contextually enriched outputs based on input sequences. Meta's \gls{LLaMA}~\cite{touvron2023llama} and OpenAI's \gls{GPT} 4~\cite{achiam2023gpt} exemplify this architecture, demonstrating exceptional performance across programming, question-answering, and language translation tasks~\cite{zhu2025overcoming}. These architectures employ autoregressive methodologies, predicting each subsequent token based on preceding tokens through \gls{CLM}. During training, the model learns to predict the next token for each position in the sequence, optimizing weights to capture contextual patterns, enabling coherent output generation.

\subsection{Evaluation Framework for Cybersecurity Domain Adaptation}
\label{sec:eval_framework}

To compare \glspl{LLM} domain adaptation in cybersecurity, we establish an evaluation framework based on five dimensions that influence practical deployment and research advancement:

\begin{itemize}[leftmargin=*]
    \item \textbf{Data Requirements.} Dataset size (tokens/words), source diversity (e.g., standards, literature, documentation), and the relationship between data scale and performance. This criterion is crucial for resource-constrained scenarios where large corpora may be unavailable.

    \item \textbf{Computational Resources.} GPU requirements, training duration, memory footprint, and costs. These factors determine both accessibility and feasibility for academic groups and smaller organizations.
    
    \item \textbf{Architecture.} Encoder-only (classification, extraction), decoder-only (generation, conversation), or encoder-decoder designs, and whether employing full continuous pretraining versus \gls{PEFT}~\cite{houlsby2019parameter} methods.
    
    \item \textbf{Domain Coverage.} Breadth and depth of cybersecurity knowledge, ranging from narrow subdomains (e.g., secure code generation, threat intelligence) to comprehensive multi-disciplinary coverage.
    
    \item \textbf{Evaluation.} Benchmark diversity and breadth (single versus multiple datasets) affecting reliability and generalizability of performance claims.
\end{itemize}

These criteria guide our comparative analysis and identify research gaps motivating our investigation.

\subsection{Related Work}
\label{sec:pre_rel_work}

We review cybersecurity-focused language model adaptation and organizing approaches by architectural paradigm.

\subsubsection{Traditional Deep Learning Approaches}

Traditional \gls{DL} methods demonstrate strong performance on specific tasks: Lee et al.~\cite{lee2019seqdroid} developed SeqDroid for Android malware detection, while Tang et al.~\cite{tang2019intrusion} employed \gls{GRU-RNN} for \gls{SDN} intrusion detection. However, these approaches lack natural language comprehension, which limits their applicability to semantic tasks such as documentation analysis or threat report generation. Recent \gls{LLM} advances address these limitations. Diaf et al.~\cite{diaf2024beyond} achieved 98\% detection accuracy combining \glspl{LLM} with \gls{LSTM} for \gls{IoT} traffic analysis, integrating both encoder and decoder architectures.

\subsubsection{Encoder-only Architectures}

Encoder-only models excel at representation and classification but are constrained in generative capabilities. Aghaei et al.~\cite{aghaei2022securebert} developed SecureBERT through \gls{BERT} adaptation with specialized tokenization, achieving superior sentiment assessment and terminology recognition compared to SecBERT~\cite{huang2024secbert}. Their subsequent SecureBERT 2.0~\cite{aghaei2025securebert} builds upon this foundation by leveraging the efficient ModernBERT architecture, which is pre-trained on a comprehensive corpus of 13.6B tokens spanning both cybersecurity text and code. Bayer et al.~\cite{bayer2024cysecbert} introduced CySecBERT, trained on diverse sources (e.g., web articles, social media, papers, vulnerability databases), explicitly validating catastrophic forgetting mitigation (a notable contribution to general capability preservation). Ranade et al.~\cite{ranade2021cybert} developed CyBERT for threat intelligence, demonstrating enhanced \gls{NER} and \gls{CKG} completion performance. While these approaches provide strong domain-specific understanding with moderate computational requirements and open-source availability, architectural constraints limit deployment to discriminative tasks.

\subsubsection{Decoder-only Architectures}

Decoder architectures enable generative applications beyond classification. He et al.~\cite{he2024instruction} proposed SafeCoder, which combines instruction tuning with security-focused fine-tuning on secure/insecure code pairs, thereby increasing StarCoder-1B's secure generation capability from 62.9\% to 92.1\%. Shestov et al.~\cite{shestov2024finetuning} fine-tuned WizardCoder~\cite{luo2023wizardcoder} for Java vulnerability detection with high \gls{ROC-AUC} and improved convergence. Fayyazi et al.~\cite{fayyazi2024advancing} augmented \gls{GPT}-3.5 with \gls{RAG} for MITRE AT\&ACK analysis, demonstrating decoder superiority over encoder-only models post-\gls{SFT}. While these approaches demonstrate strong performance with modest computational requirements (1B-scale models) or no retraining (\gls{RAG}), they either target narrow domains (e.g., code security) or rely on proprietary models, thereby limiting reproducibility.

\subsubsection{Cybersecurity Specialization}

Yu et al.~\cite{yu2025primus} introduced Llama-Primus-Base, the most comparable work to our investigation, which employs a similar \gls{DAP} methodology using 2.77 billion tokens from Primus-FineWeb. The model achieves 0.667 (CTI-MCQ), 0.866 (CyberMetric), and 0.500 (SecEval) accuracy under 5-shot evaluation with full open-source availability. Similarly, Kassianik et al.~\cite{kassianik2025llama} developed Llama-3.1-FoundationAI-SecurityLLM-Base-8B through continued pretraining on 5 billion tokens, achieving scores of 0.662 (CTI-MCQ) and 0.848 (CyberMetric) under a 5-shot evaluation. Both approaches enable comprehensive coverage but raise questions about data efficiency, specifically whether equivalent specialization can be achieved through more focused curation.

\subsubsection{Research Gap}

\textbf{Data Efficiency.} Existing comprehensive approaches employ massive datasets (Llama-Primus: 2.77B tokens; FoundationAI: 5B tokens), while efficient approaches target narrow subdomains. Our work investigates whether broad adaptation is achievable with 118.8 million tokens (a 23-fold reduction) while maintaining comprehensive coverage across standards, literature, and documentation.

\begin{figure*}[!htbp]
    \centering
    \includegraphics[width=0.85\textwidth]{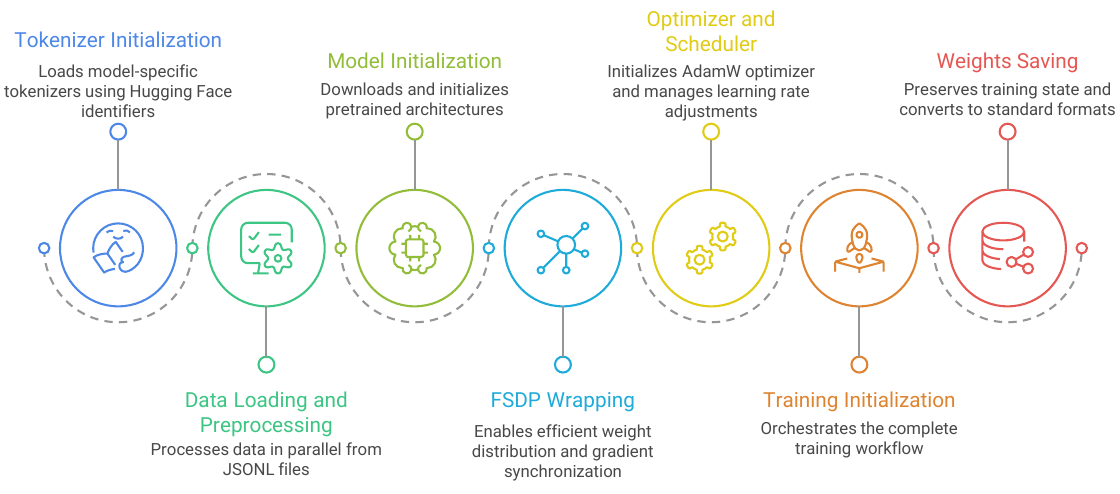}
    \caption{\textcolor{customblue}{Our domain-adaptive continuous pretraining pipeline. The workflow illustrates the systematic transformation from raw cybersecurity corpus collection through preprocessing, JSONL formatting, and distributed training to produce specialized domain-adapted model weights.}}
    \label{fig:dap_workflow}
\end{figure*}

\textbf{Scale-Dependent Learning Dynamics.} Limited investigation exists into adaptation effectiveness across model scales under comparable training regimes. Our systematic evaluation across three architectures (Llama-3.1-8B, DeepSeek-R1-Distill-Qwen-14B, Llama-3.3-70B-Instruct) provides empirical insights into optimal resource allocation strategies.

\textbf{Catastrophic Forgetting.} While CySecBERT explicitly validated general capability preservation, most domain adaptation research does not address any potential degradation of foundational language understanding. Our conservative training approach—employing low learning rates (1×10$^{-6}$), limited epochs (2-3), and frozen embeddings—was specifically designed to mitigate this risk, though comprehensive validation on general benchmarks remains a direction for future work due to resource constraints (as discussed in Section~\ref{sec:discussion}).

Our positioning within this landscape emphasizes practical efficiency, achieving competitive cybersecurity specialization with substantially reduced data requirements while maintaining decoder-only generative capabilities essential for tasks such as threat analysis, vulnerability assessment, and security documentation generation.

%% file: Sections/3_methodology.tex
\section{Methodology}
\label{sec:methodology}
This section presents our proposed \glspl{LLM} \gls{DAP} methodology for cybersecurity applications. Our end-to-end pipeline is depicted in Figure~\ref{fig:dap_workflow}, which illustrates the systematic workflow comprising seven integrated components that transform raw cybersecurity data into specialized domain-adapted model weights.

The pipeline orchestrates three primary phases: \textit{initialization}, \textit{distributed training}, and \textit{persistence}. The initialization phase establishes the foundation by loading pretrained model architectures and their associated tokenizers, followed by preprocessing domain-specific data into standardized formats suitable for parallel processing. The distributed training phase implements \gls{FSDP} to enable efficient weight distribution and gradient synchronization across GPU clusters, addressing the computational demands of large-scale language models. Training orchestration manages the iterative optimization process, including forward propagation, loss computation, gradient backpropagation, and periodic checkpointing for fault tolerance and reliability. The persistence phase preserves the complete training state and converts adapted weights into formats compatible with downstream evaluation and deployment.

This modular pipeline architecture enables scalable domain adaptation while maintaining reproducibility and computational efficiency across varying model scales and hardware configurations.

\subsection{Methodological Rationale}
\label{sec:method_rationale}
Our investigation employs \gls{DAP} rather than adapter-based approaches such as \gls{PEFT}~\cite{houlsby2019parameter} or \gls{LoRA}~\cite{hu2021lora} to achieve a fundamentally different objective: instilling an intrinsic understanding of cybersecurity within the model's foundational representations. While adapter-based methods demonstrate strong performance for task-specific optimization (e.g., vulnerability classification, log parsing), they introduce specialized parameters on top of frozen base models without modifying the underlying domain comprehension. This architectural constraint limits their capacity to embed deep semantic understanding and domain-specific reasoning patterns into the model's core representations.

\gls{DAP}, conversely, enables comprehensive knowledge integration through continued pretraining on domain-specific corpora, allowing the model to internalize cybersecurity terminology, conceptual relationships, and contextual patterns at the representational level. This foundational enhancement proves particularly valuable for complex reasoning tasks that require nuanced domain understanding, such as threat analysis, security documentation generation, and multi-step vulnerability assessments. The resulting domain-adapted models subsequently serve as robust foundations for downstream specialization through instruction tuning, \gls{PEFT}, or task-specific fine-tuning, enabling efficient adaptation to specific applications including intrusion detection, secure code generation, and compliance analysis.

Our methodology thus prioritizes building foundational cybersecurity-aware language models that possess inherent domain expertise, creating versatile bases for subsequent specialization across diverse cybersecurity applications rather than optimizing for any single narrow task.

\subsection{Dataset Preparation}
The foundation of our \gls{DAP} methodology rests upon a meticulously curated cybersecurity corpus that achieves an optimal balance between domain specificity and comprehensive coverage. In contrast to pretraining datasets that encompass petabyte-scale collections, our domain adaptation strategy employs a focused dataset specifically designed to instill domain-specific terminologies and contextual nuances. {\color{customblue} This deliberate constraint in dataset scope serves dual purposes: ensuring sufficient exposure to cybersecurity concepts while aiming to reduce the risk of the model overspecializing and losing its general language understanding capabilities, though preservation of general capabilities was not empirically validated in this work}.

\textbf{Data Curation.} We developed a custom cybersecurity corpus through the systematic collection of materials from three primary categories. Standards and regulations form the cornerstone of our dataset, incorporating authoritative documents from recognized bodies such as NIST and ENISA. These regulatory frameworks provide a solid foundation in formal cybersecurity terminology and compliance requirements. Research papers sourced from academic platforms, including arXiv, contribute cutting-edge insights and emerging threat landscapes to our corpus. Technical literature, including books on network security, malware analysis, and related domains, completes our dataset, ensuring comprehensive coverage of both theoretical foundations and practical applications. This targeted curation approach ensures a thorough representation of cybersecurity subdomains while maintaining a dataset scope suitable for domain adaptation rather than full model retraining.

\textbf{Data Processing Pipeline.} Raw documents undergo systematic preprocessing through a pipeline designed to maximize data quality while preserving contextual integrity. The process begins by extracting text from PDF and text files, then converting it to JSON for structured representation. Subsequently, we apply \gls{RegEx} patterns to comprehensively remove noise, eliminating extraneous elements such as page numbers, citations, and empty lines that could interfere with the learning process. The cleaned data is then transformed into JSONL format, ensuring compatibility with modern training frameworks. Throughout this process, we maintain paragraph-level segmentation to preserve contextual coherence, recognizing that cybersecurity concepts often require surrounding context for proper interpretation. Notably, no manual labeling is required, as models are trained unsupervised via \gls{CLM}, significantly reducing preprocessing overhead while maintaining training effectiveness.

\subsection{Model Selection and Architecture}
Our methodology employs decoder-only architectures, chosen for their demonstrated performance in autoregressive text generation tasks compared to encoder-only or encoder-decoder alternatives. This architectural decision stems from decoder models' streamlined approach to next-token prediction, which aligns naturally with our objective of generating contextually appropriate cybersecurity responses. The selection process prioritizes three critical criteria that balance practical constraints with performance requirements. Computational efficiency remains paramount given resource constraints typical in research environments, necessitating models that deliver optimal performance without requiring extensive \gls{GPU} clusters. Finally, we consider the ratio of pretraining corpus size to model parameters, ensuring that the selected models possess a sufficient knowledge base to support effective domain adaptation.

Initial experiments utilize \gls{LLaMA} 3.2 variants with 1B and 3B parameters, models pretrained on approximately 9 trillion tokens. This extensive pretraining provides a substantial knowledge foundation while maintaining computational feasibility for domain adaptation. Subsequent experiments scale to larger models, such as 8B-parameter models, allowing a systematic evaluation of the performance improvements achievable through increased model capacity. This progressive scaling strategy enables empirical determination of the optimal trade-off between computational requirements and domain adaptation effectiveness.

\subsection{Domain-Adaptive Training Framework}

Our training methodology leverages distributed multi-\gls{GPU} processing to accommodate the computational demands of modern \glspl{LLM} while implementing carefully calibrated strategies to preserve general knowledge during domain specialization.

\textbf{Training Configuration.} The configuration begins with tokenization using pretrained tokenizers without modification, based on the reasonable assumption that tokenizers trained on trillions of tokens possess comprehensive vocabulary coverage for our domain-specific corpus. This approach avoids potential vocabulary fragmentation that could arise from retraining tokenizers on limited domain data. Specifically, we freeze the embedding layer throughout training to preserve learned token representations acquired during pretraining. This strategy prevents degradation of fundamental language understanding while allowing higher layers to adapt to domain-specific patterns.

Learning rate selection requires particular attention in domain adaptation scenarios. {\color{customblue} We deliberately employ small learning rates to facilitate gradual weight adjustments, with the intention of mitigating catastrophic forgetting that could result from aggressive parameter updates, although this preservation has not been empirically validated. Similarly, we limit training to a minimal number of epochs, typically one to three complete passes through the dataset, thereby aiming to avoid overfitting on domain-specific patterns that could compromise generalization capability.}

\textbf{Training Process.} The domain adaptation process follows a systematic protocol designed for reproducibility and optimal results. Training commences with the initialization of pre-trained model weights, establishing the foundation of general language understanding upon which domain knowledge is built. Following initialization, we freeze the embedding layer to maintain vocabulary representations while allowing other parameters to adapt. Training proceeds with our carefully controlled \gls{LR} schedule, ensuring gradual weight adjustments that preserve existing knowledge while incorporating domain expertise. Upon completion, adapted model weights are saved in appropriate formats such as binary or safe tensors, facilitating efficient deployment and further experimentation.

\subsection{Evaluation Framework}
\label{sec:eval_framework}

Our comprehensive evaluation methodology includes both training-phase metrics and post-training benchmarks, providing a multifaceted assessment of domain adaptation effectiveness across various dimensions of model performance.

\textbf{Training Phase Evaluation.} During \gls{DAP}, we continuously monitor perplexity as the primary optimization metric, calculated as $\text{PPL} = \exp\left(-\frac{1}{N} \sum_{i=1}^N \log_2 q(x_i)\right)$, where $N$ represents the total number of tokens in the sequence and $q(x_i)$ denotes the probability assigned by the model to token $x_i$. Lower perplexity values indicate higher model confidence in next-token prediction, serving as a proxy for improved domain understanding. However, we recognize that perplexity alone cannot fully assess contextual comprehension or practical applicability, necessitating comprehensive post-training evaluation.

\textbf{Post-Training Assessment.} Domain-adapted models undergo rigorous evaluation using established cybersecurity benchmarks to quantify improvement in domain-specific understanding. Performance assessment on datasets such as CyberMetric~\cite{tihanyi2024cybermetric} employs \gls{MCQ} format questions that test various aspects of cybersecurity knowledge. We conduct direct accuracy comparisons between original and adapted models on identical evaluation sets, providing clear quantitative measures of domain adaptation effectiveness. 

%% file: Sections/4_implementaion.tex
\section{Implementation}
\label{sec:implementation}

The \gls{DAP} of \glspl{LLM} presents substantial computational challenges due to their massive parameter counts and quadratic attention complexity $O(n^2)$, where $n$ represents the prompt length. For instance, \gls{GPT}-2 employs 768-dimensional vectors per token to capture embedding and positional information, resulting in prohibitive resource requirements that necessitate distributed training environments with weight sharding and aggregation mechanisms. Our training implementation employs a mixed-precision strategy to optimize computational efficiency while preserving numerical stability. Models are loaded in half-precision formats (FP16/BF16) to reduce memory footprint, while most computations are performed in half-precision for enhanced throughput. Critical operations, including loss calculation, utilize full-precision (FP32) with gradient scaling to prevent numerical instability and gradient underflow that would otherwise compromise training convergence.

This section presents a comprehensive implementation framework for a cybersecurity-focused DAP that utilizes AWS SageMaker's P5 instances~\cite{amazonSagemaker} equipped with 8 NVIDIA H100 GPUs, addressing both technical constraints and domain-specific requirements.

\subsection{Dataset Construction and Preprocessing}
\label{sec:dataset_overview}

The absence of comprehensive cybersecurity-focused datasets led to the construction of a custom corpus comprising 126 million words (132 million tokens using \gls{LLaMA} tokenizer). This corpus integrates diverse sources to ensure comprehensive domain coverage while maintaining linguistic variety essential for effective model adaptation.

\textbf{Dataset Coverage.} Our curated dataset encompasses three primary categories, each contributing unique contextual information: (i) \textit{Standards \& Regulations:} Industry-recognized documents from organizations such as ISO provide formal compliance frameworks and security guidelines. These documents enable the model to extract contextual information and generate responses adhering to established standards, particularly valuable for compliance analysis during \gls{SFT} for downstream tasks. (ii) \textit{Papers:} Academic publications encompass cutting-edge cybersecurity research, providing formal, well-structured, fact-based content. This exposure enables the model to learn academic writing styles, facilitating the generation of concise summaries and technical reports grounded in sound reasoning and logic. (iii) \textit{Books:} Technical literature presents concepts at varying complexity levels with extended context lengths. This diversity enables the model to develop longer contextual memory while learning to generate user-friendly responses across different writing styles, from descriptive to explanatory formats.

During our experimental phase, the Primus-Seed and Primus-FineWeb~\cite{yu2025primus} datasets became available. While Primus-FineWeb's 2.57 billion tokens posed risks of overfitting, the expert-curated Primus-Seed (0.2 billion tokens) dataset, containing content from reputable sources including MITRE, was preserved for potential integration in subsequent fine-tuning stages or downstream task optimization.

\subsubsection{Pre-processing} The heterogeneous nature of our curated documents required comprehensive preprocessing to eliminate noise and standardize formats. Our pipeline addresses three critical aspects:

\textbf{Data Filtering.} Documents collected from sources, including the NIST Open Document Library and arXiv repositories, underwent multi-stage filtering. Manual curation combined with AI-based classification addressed the challenge of ambiguous domain categorization, particularly for research papers with multiple keywords. The classification prompt specifically evaluated strict adherence to cybersecurity or its subdomains. \gls{RegEx} filtering eliminated patterns already learned during pretraining (i.e., email formats, phone numbers, and references), maintaining minimal noise-to-content ratios to prevent memorization of irrelevant patterns.

\begin{figure}[H]
    \centering
    \includegraphics[width=0.8\linewidth]{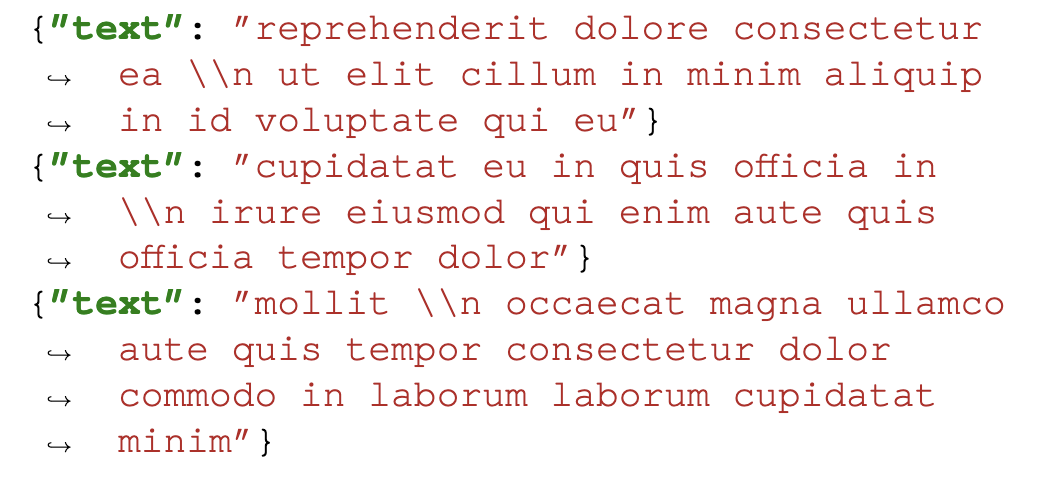}
    \caption{Sample \gls{JSONL} structure.}
    \label{fig:jsonl_sample}
\end{figure}

\textbf{Data Formatting.} The multi-format corpus (PDFs, text files, structured and unstructured content) required standardization for tokenization. Text extraction and filtering preceded segmentation into fixed sequence lengths, minimizing padding overhead during tokenization. The processed content was formatted as \gls{JSONL} entries (depicted in Figure~\ref{fig:jsonl_sample}), facilitating efficient data loading.

\begin{figure}[H]
    \centering
    \includegraphics[width=0.8\linewidth]{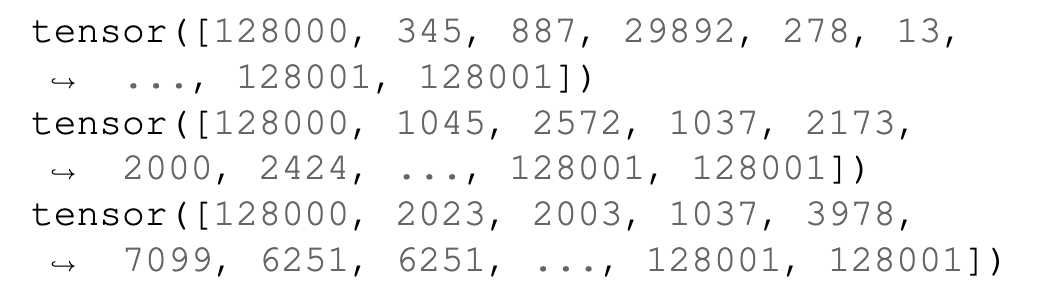}
    \caption{Sample tokenized input IDs (\gls{LLaMA}3 tokenizer).}
    \label{fig:llama_input_tensors}
\end{figure}

Context window constraints of \glspl{LLM} necessitated sequence truncation to specified lengths (1024 or 2048 tokens), as illustrated in Figure~\ref{fig:llama_input_tensors}. This approach simplifies context retention across sequences by eliminating random splits.

\textbf{Data Splitting.} The dataset allocation follows a 90-10 train-test split, with the training portion further stratified into three variants with $\approx 1 \text{ million}$, $\approx50\text{ million}$, and $\approx118.8$ million tokens, respectively. Random shuffling during variant creation prevents ordering bias, ensuring robust learning across different model scales.

\subsection{Domain Adaptive Continuous Pretraining Pipeline}
\label{sec:dap_pipeline}

\textbf{Model Selection.} Model selection balanced multiple constraints: architectural diversity, open-source availability, and computational feasibility. While state-of-the-art models like \gls{GPT}-4 and Claude 3.7~\cite{anthropicClaude37} remain proprietary, we selected open-source alternatives demonstrating strong performance across mathematical, linguistic, and reasoning tasks. Initial pipeline validation utilized \gls{LLaMA}-3.2-3B before proceeding to larger architectures. Table~\ref{tab:models_to_use} presents the final model selection with corresponding dataset allocations, designed to evaluate scale-dependent learning dynamics.

\begin{table}[!ht]
\centering
\caption{Approximate training‑data size for each model.}
\label{tab:models_to_use}
\resizebox{\columnwidth}{!}{%
\begin{tabular}{|l|c|}
\hline
\textbf{Model name} & \textbf{Approximate dataset size in million [$\times10^{6}$] tokens} \\
\hline
Llama-3.1-8B & 1 \\
DeepSeek‑R1‑Distill‑Qwen‑14B (DSR1D-Qwen-14B) & 50 \\
Llama-3.3-70B-Instruct (Llama-3.3-70B-Ins) & 118.8 \\
\hline
\end{tabular}%
}
\end{table}

Resource requirements for model loading follows: $\text{Model Size} = 4N\times10^9 \,\text{Bytes}$, where $N \in \{1, 3, 8, 13, 14,$ 
$ 32, 70, 671, \dots \}$ represents billions of parameters. Full training memory requirements approximate:

{\footnotesize\begin{equation}
    \text{Memory Requirement} \approx \alpha \times \text{Model Size}, 
    \quad \text{where } \alpha \in [3, 5]
\label{eq:full_mem_req}
\end{equation}}

This scaling factor accounts for gradient storage, activations, and optimizer states during forward and backward passes. While half-precision formats (\textit{float16} or \textit{bfloat16}~\cite{mediumDifferenceBetweenFloat16AndBfloat16}) reduce memory requirements by 50\%, models like DeepSeek-R1-671B remain computationally prohibitive given current resource constraints.

\subsubsection{Hyperparameters Configuration}

{\color{customblue} Hyperparameter selection influences learning dynamics, particularly for \glspl{LLM}, where the massive number of parameters amplifies the risks of overfitting. Our configuration below prioritizes stability while enabling effective domain knowledge acquisition.}

\textbf{Learning Rate.} Configured at $\approx 1 \times 10^{-6}$, this conservative value prevents overfitting while enabling gradual adaptation to domain-specific patterns. This selection balances the competing objectives of knowledge acquisition and retention.

\textbf{Maximum Sequence Length.} Constrained to $\{1024, 2048\}$ tokens due to memory limitations in attention computation. Longer sequences enhance contextual understanding through self-attention mechanisms but proportionally increase memory consumption for attention matrices, activations, and gradients.

\textbf{Batch Size.} Per-device batch sizes of $\{1, 2\}$ accommodate GPU memory constraints. Distributed training across 4-8 GPU instances achieves effective parallelism. For instance, 4 compute instances with 4 GPUs each yield an effective batch size of $1\times4\times4 = 16$.

\textbf{Gradient Accumulation Step.} Mini-batch processing with gradient accumulation maintains computational efficiency while achieving effects comparable to larger batch sizes. Total effective batch size is calculated as $\text{Total Batch Size} = \text{MBS} \times \text{GAS} \times \text{World Size}$. Where MBS denotes per-device mini-batch size, GAS represents gradient accumulation steps, and World Size is defined as $\text{World Size} = \text{Total Devices} \times \text{GPUs per Device}$.

Additional configurations include (i) maximum 2-3 training epochs to prevent domain overfitting, (ii) cosine learning rate scheduling with warm-up to prevent exploding gradients, (iii) and AdamW optimizer~\cite{loshchilov2017decoupled} providing stable weight updates through decoupled weight decay regularization.

\subsection{Model Fine-tuning}
The distributed training pipeline addresses the computational infeasibility of single-device \gls{LLM} training through systematic parallelization strategies. Our implementation leverages PyTorch's \gls{FSDP} framework, combining the benefits of data and model parallelism through weight sharding with distributed batch processing. The implementation comprises the following key components:
\begin{enumerate}[leftmargin=*]
    \item \textbf{Tokenizer Initialization.} The AutoTokenizer API from the Transformers library loads model-specific tokenizers via Hugging Face identifiers (e.g., \textit{meta-llama/Llama-3.1-8B}) or local storage paths for pre-saved configurations.

    \item \textbf{Data Loading and Preprocessing.} PyTorch's DistributedSampler and DataLoader utilities enable parallel data processing from pre-formatted \gls{JSONL} files. The pipeline generates identical input-output sequence pairs for unsupervised \gls{DAP}, returning properly formatted tensor batches.

    \item \textbf{Model Initialization.} AutoModelForCausalLM downloads and initializes pretrained architectures from Hugging Face Hub, automatically managing weight shard distribution across available devices.

    \item \textbf{\gls{FSDP} Wrapping.} The FullyShardedDataParallel API enables efficient weight distribution and gradient synchronization. This approach supersedes the Transformers Trainer utility due to enhanced customization capabilities for our specific requirements.

    \item \textbf{Optimizer and Scheduler.} AdamW optimizer initialization via PyTorch's optim utility receives model parameters for weight updates. The learning rate scheduler manages adaptive rate adjustments throughout training phases.

    \item \textbf{Training Initialization.} A custom Trainer class orchestrates the complete training workflow: dataset iteration, forward pass computation, loss calculation, gradient backpropagation, weight updates, and checkpoint persistence. Critical recovery capabilities store current step and world size information, enabling mid-epoch training resumption after interruptions.

    \item \textbf{Weights Saving.} Checkpoint management preserves a comprehensive training state, including model weights, optimizer states, and scheduler configurations, using PyTorch's distributed API in sharded format. Post-training conversion via Transformers' \textit{save\_pretrained} API produces standard formats compatible with \textit{from\_pretrained} or \textit{AutoModelForCausalLM} loading mechanisms, facilitating seamless downstream evaluation using utilities like \textit{generate}.

\end{enumerate}

%% file: Sections/5_performance_evaluation.tex
\section{Performance Evaluation}
\label{sec:perf_eval}
The evaluation of domain-adapted \glspl{LLM} requires substantial computational resources due to their scale and complexity. For instance, loading the \textit{Llama-3.1-8B} model in half-precision requires approximately 29.9 GB of memory, calculated as 
$\text{GPU Memory Required} > 8 \times 10^9 \times 2 \ \text{bytes} \approx 29.9 \ \text{GB}$. Larger models demand proportionally greater resources, necessitating cloud-based evaluation infrastructure. Although inference requirements are less stringent than domain adaptation, the \gls{GPU} must simultaneously accommodate model parameters, tokenized inputs, and intermediate computations, making traditional \gls{CPU}-based evaluation infeasible without quantization\footnote{Quantization can lead to performance degradation.}. 

\subsection{Experimental Setup}

\begin{figure}[!h]
    \centering
    \includegraphics[width=\linewidth]{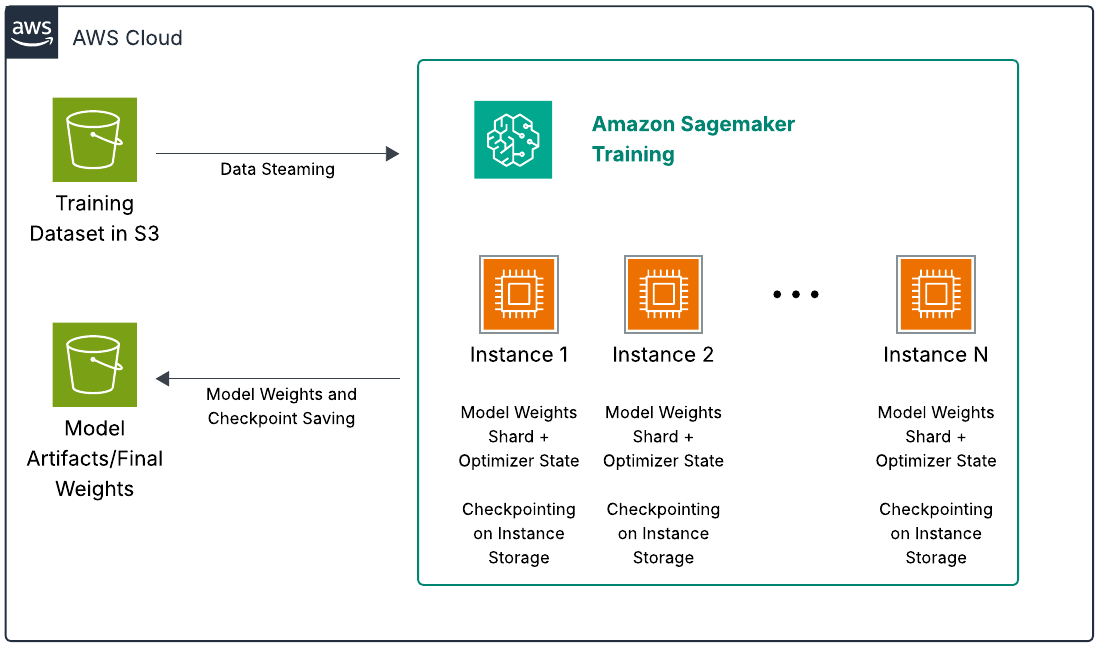}
    \caption{AWS reference architecture for distributed training and evaluation infrastructure.}
    \label{fig:aws_ref_arch_diagram}
\end{figure}

{\color{customblue}
Our distributed training and evaluation pipeline leverages the AWS reference architecture, illustrated in Figure~\ref{fig:aws_ref_arch_diagram}, which is orchestrated through Amazon SageMaker to manage computational resources and coordinate data flow. The training dataset is stored persistently in an AWS S3 bucket and streamed directly to the SageMaker training cluster during adaptation. We employ a distributed data-parallel strategy, in which SageMaker provisions multiple instances (Instance 1 through Instance N) that run in parallel. Consistent with the \gls{FSDP} methodology detailed in Section~\ref{sec:implementation}, model parameters (Model Weights Shard) and optimizer states are sharded across all available \glspl{GPU} on these instances. Training checkpoints are saved to instance storage for efficient state recovery in case of interruptions. Upon completion, the final model artifacts, including the adapted weights, are aggregated and stored in a specific S3 bucket for subsequent evaluation and deployment.
}

For inference evaluation, our pipeline uses AWS SageMaker inference endpoints with mixed-precision configurations (\textit{FP32}/\textit{FP16}/\textit{BF16}) during \gls{DAP}, and half-precision formats for memory-optimized evaluation. Table~\ref{tab:model_instance_mapping} presents the instance allocation strategy for each \gls{LLM} architecture.

\begin{table}[!h]
\centering
\caption{Model to Instance Mapping (Inference)}
\label{tab:model_instance_mapping}
\resizebox{0.5\columnwidth}{!}{%
\begin{tabular}{|l|l|}
\hline
\textbf{Model} & \textbf{Instance} \\
\hline
Llama-3.1-8B & \textit{ml.g5.12xlarge} \\
DSR1D-Qwen-14B & \textit{ml.g5.12xlarge} \\
Llama-3.3-70B-Ins & \textit{ml.p4d.24xlarge} \\
\hline
\end{tabular}%
}
\end{table}

The \textit{ml.g5.12xlarge} instances utilize NVIDIA A10G GPUs (24 GiB of memory per GPU, totaling 96 GiB), which are sufficient for 8B and 14B models in half-precision. The 70B model requires \textit{ml.p4d.24xlarge} instances with NVIDIA A100 GPUs, optimized for \textit{bfloat16} operations and providing enhanced computational capacity. Table~\ref{tab:aws_instance_specs} details the hardware specifications. The theoretical compute capacity per instance is computed as $\text{Compute Capacity} = \text{Num of GPUs} \times \text{TFLOPS per GPU}$. Our implementation utilizes the SageMaker SDK with HuggingFace Transformers (version 4.48.0), PyTorch 2.3.0, and Python 3.11, enabling direct model deployment and evaluation through the established \glspl{API}.

\begin{table}[!h]
\centering
\caption{AWS instance specifications~\cite{awsSageMakerPricing}. (OD: On-Demand)}
\label{tab:aws_instance_specs}
\resizebox{\columnwidth}{!}{%
\begin{tabular}{|l|c|c|c|c|c|}
\hline
\textbf{Instance} & \textbf{\# GPUs} & \textbf{GPU Type} & \textbf{Mem/GPU (GiB)} & \textbf{TFLOPS (FP16)} & \textbf{(OD) Price/hr} \\
\hline
\textit{ml.g5.12xlarge} & 4 & NVIDIA A10G & 24 & 31.52 & \$7.09 \\*
\textit{ml.p4d.24xlarge} & 8 & NVIDIA A100 & 40 & 19.5 & \$25.251 \\
\hline
\end{tabular}%
}
\end{table}

{\color{black} \textbf{Baseline Configuration Parameters.} To validate the efficacy of full-parameter adaptation, we compared our approach against a \gls{PEFT} baseline using \gls{LoRA}. The \gls{LoRA} configuration utilized a rank of $r=16$, $\alpha=32$, and dropout of $0.05$, targeting the query, key, value, and output projection modules within the attention mechanism.}

{\color{customblue}
\subsection{Decoding Configuration}
\label{sec:decoding_config}

To ensure reproducible and deterministic evaluation across all benchmarks, we employ greedy decoding with fixed parameters: \texttt{do\_sample = False} and \texttt{temperature = 0}. This configuration eliminates stochastic sampling and probability scaling, producing a single deterministic output per prompt. By removing decoding randomness, our evaluation isolates model behavior and domain knowledge acquisition from variability introduced by sampling strategies, enabling consistent cross-benchmark comparisons. All reported results reflect single-pass evaluations under this standardized configuration.
}

\subsection{Evaluation Benchmarks}
We assess cybersecurity domain adaptation using three specialized datasets that comprehensively evaluate different aspects of cybersecurity knowledge. The CTI-MCQ dataset~\cite{alam2024ctibench} comprises 2500 multiple-choice questions specifically designed to evaluate \gls{CTI} understanding, testing models' comprehension of threat identification, analysis methodologies, and intelligence-driven security practices. The CyberMetric benchmark~\cite{tihanyi2024cybermetric} comprises 2000 multiple-choice questions that assess general knowledge across various cybersecurity domains, offering comprehensive coverage of fundamental security principles, best practices, and technical competencies. Finally, the SecEval dataset~\cite{githubSecEval} comprises over 2000 multiple-choice questions that focus on broad cybersecurity knowledge evaluation, spanning various security subdomains, including vulnerability assessment, risk management, and defensive strategies.

These benchmarks span cybersecurity standards, network and cloud security, cryptography, identity and access management, and threat mitigation strategies. Due to computational constraints, we evaluated the 70B model on reduced samples for CyberMetric (1000 questions) while maintaining full evaluation sets for other models and datasets.

\subsection{Results and Analysis}
We evaluate the model performance using standard accuracy metrics, i.e., $\text{Accuracy} = \frac{1}{N} \sum_{i=1}^{N} \text{1}(y_i = \hat{y}_i)$. Here $y_i$ and $\hat{y}_i$ are the ground truth and predictions from the models, respectively. ${N}$ is the total data samples in the evaluation dataset. If the prediction and ground truth are the same $\text{1}(y_i = \hat{y}_i)$ evaluate to $1$ otherwise $0$. For SecEval, which permits multiple correct answers, we employ the Jaccard Index for per-question accuracy, i.e.,  $\text{Jaccard Index}(GT, Pred) = \frac{|GT \cap Pred|}{|GT \cup Pred|}$, where $GT$ is the ground truth set and $Pred$ is the model response (extracted) answer set.

\begin{figure}[!h]
    \centering
    \includegraphics[width=0.8\linewidth]{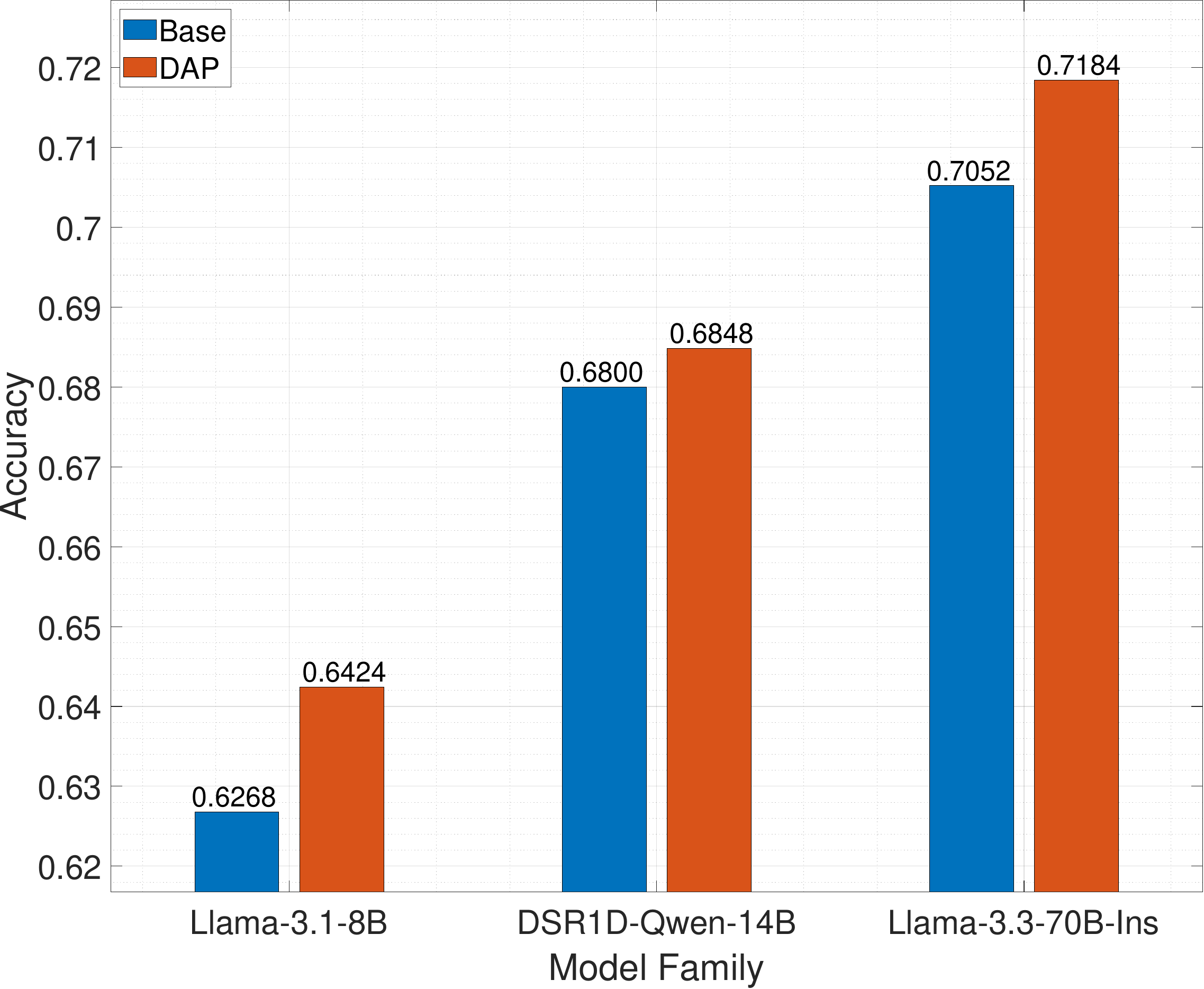}
    \caption{Performance comparison of base and domain-adapted models on CTI-MCQ benchmark. Domain adaptation yields consistent improvements across all architectures, with accuracy gains of 1.6\% for Llama-3.1-8B, 0.5\% for DSR1D-Qwen-14B, and 1.3\% for Llama-3.3-70B-Ins.}
    \label{fig:acc_cmp_ctimcq}
\end{figure}

\begin{figure}[!h]
    \centering
    \includegraphics[width=0.8\linewidth]{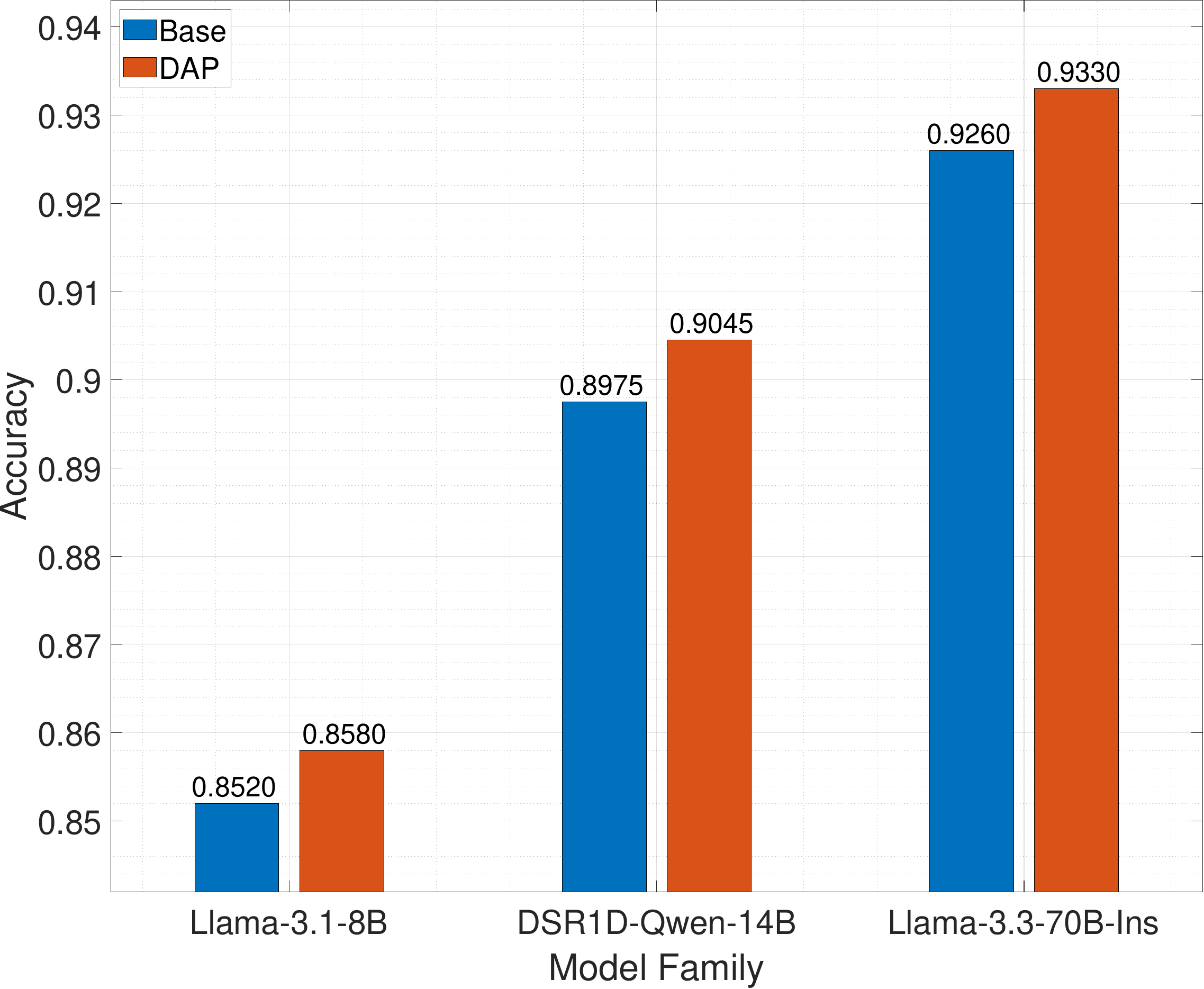}
    \caption{Performance comparison of base and domain-adapted models on CyberMetric (2000) benchmark. DAP consistently improves cybersecurity knowledge across all \glspl{LLM}, with accuracy gains ranging from 0.6\% to 0.7\%.}
    \label{fig:acc_cmp_cybermetric}
\end{figure}

\begin{figure}[!h]
    \centering
    \includegraphics[width=0.8\linewidth]{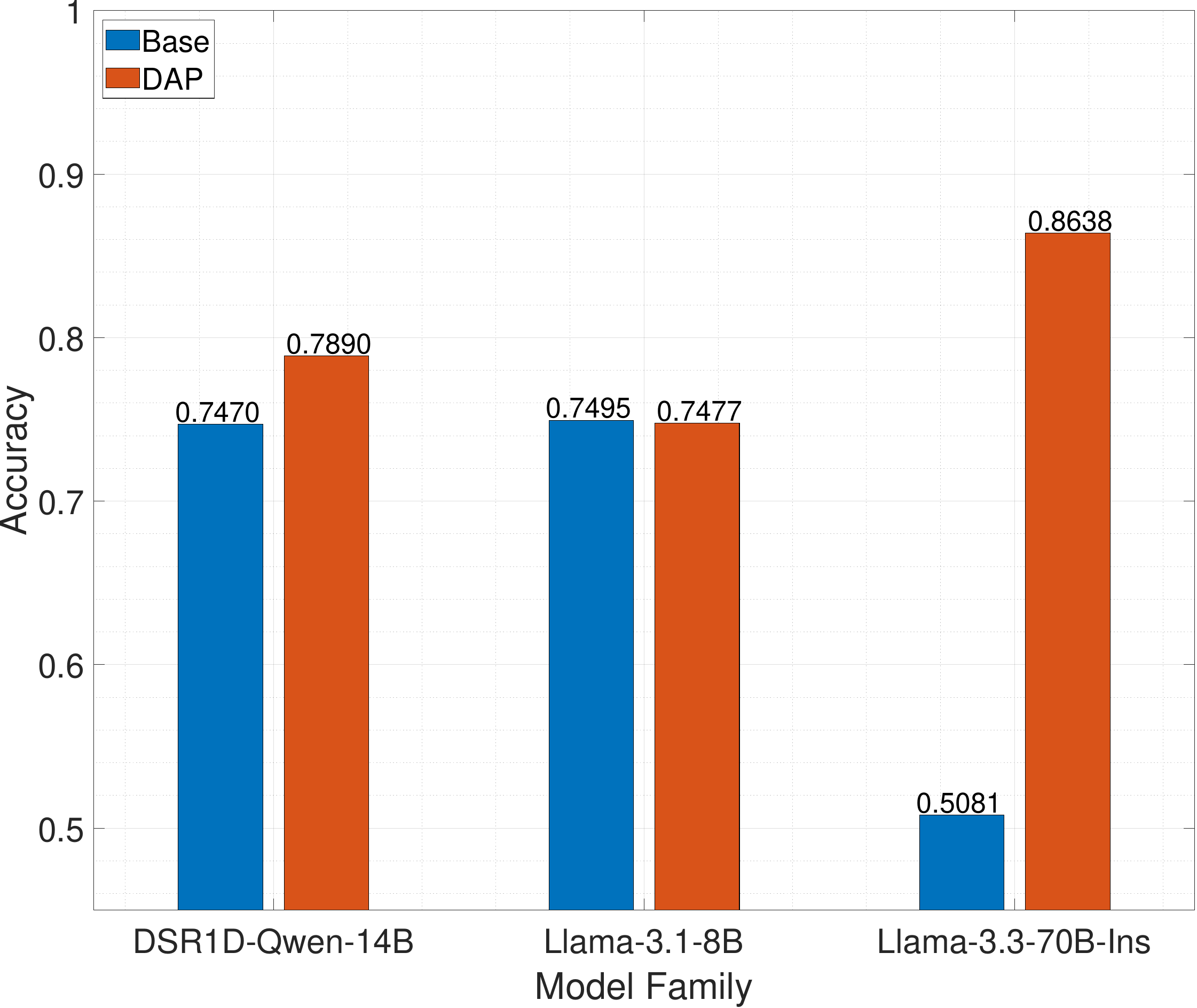}
    \caption{Performance comparison of base and domain-adapted models on the SecEval benchmark. Domain adaptation demonstrates significant improvements for DSR1D-Qwen-14B (0.747 to 0.789) and Llama-3.3-70B-Ins (0.508 to 0.864), while Llama-3.1-8B shows minimal change.}
    \label{fig:acc_cmp_seceval}
\end{figure}

\textbf{Performance on CTI-MCQ.} Figure~\ref{fig:acc_cmp_ctimcq} demonstrates the effectiveness of domain-adaptive continuous pretraining across all evaluated model architectures on the CTI-MCQ benchmark, which specifically assesses Cyber Threat Intelligence understanding. The results reveal consistent performance improvements following domain adaptation, with notable variations in the magnitude of improvement across different model scales. The Llama-3.1-8B model exhibits the most substantial relative improvement, achieving a 1.6\% accuracy gain from 0.6268 to 0.6424, representing a meaningful enhancement in cybersecurity knowledge acquisition despite the model's smaller parameter count. The DSR1D-Qwen-14B architecture demonstrates more modest but consistent improvement with a 0.5\% increase from 0.6800 to 0.6848. In contrast, the largest model, Llama-3.3-70B-Ins, achieves a 1.3\% improvement, rising from 0.7052 to 0.7184, and ultimately attains the highest performance across all evaluated architectures. These results validate the efficacy of our domain adaptation methodology in enhancing specialized cybersecurity comprehension while maintaining architectural diversity in performance gains.

\textbf{Performance on CyberMetric.} Figure~\ref{fig:acc_cmp_cybermetric} illustrates the performance characteristics of our domain-adapted models on the CyberMetric benchmark, which evaluates general cybersecurity knowledge across diverse security domains. The results demonstrate remarkably consistent improvement patterns, with all three architectures achieving accuracy gains within a range of 0.6\% to 0.7\%. Specifically, the Llama-3.1-8B model improves from 0.8520 to 0.8580, while the DSR1D-Qwen-14B achieves enhancement from 0.8975 to 0.9045, and the Llama-3.3-70B-Ins advances from 0.9260 to 0.9330. These improvements across different model scales imply that our domain adaptation approach effectively instills general cybersecurity knowledge regardless of architectural complexity. Notably, all models achieve relatively high baseline performance on this benchmark, with the domain adaptation process providing incremental yet consistent enhancements that validate the robustness of our training methodology across diverse cybersecurity knowledge domains.

\textbf{Performance on SecEval.} Figure~\ref{fig:acc_cmp_seceval} presents the most different results across architectures within our evaluation framework, revealing complex interactions between baseline model capabilities and domain adaptation efficacy on the SecEval benchmark. This evaluation dataset employs a sophisticated multi-answer format utilizing Jaccard Index scoring, which demands advanced instruction-following and contextual reasoning capabilities. The results demonstrate three distinct adaptation patterns that highlight fundamental relationships between pre-existing model alignment and domain specialization potential.

The DSR1D-Qwen-14B architecture demonstrates moderate yet meaningful progression from 0.7470 to 0.7890, reflecting its intermediate position between raw pretraining and full instruction alignment, with knowledge distillation foundations providing a stable platform for domain enhancement. In contrast, the Llama-3.1-8B model demonstrates minimal variation, i.e, $\approx$ 0.002 difference, between the base and DAP model. This indicates that this intermediate-scale architecture maintains a robust baseline performance that proves relatively resistant to further domain-specific enhancements, which could be attributed to the relatively small size of the \gls{DAP} dataset, i.e., 1 million tokens, for an \gls{LLM} with 8 billion parameters.

Remarkably, the Llama-3.3-70B-Ins model exhibits a significant improvement, increasing from 0.5081 to 0.8638, representing a 70\% relative increase, which is the most substantial performance gain observed across our entire evaluation framework. This enhancement reveals that despite the model's extensive parameter count and sophisticated pre-training, the baseline architecture exhibited significant deficiencies in complex multi-answer reasoning tasks, which domain adaptation effectively addressed. 

\textit{Key Takeaway.} The aforementioned outcomes challenge conventional assumptions about model scale, training data requirements, and inherent capabilities. Our findings demonstrate that even large, instruction-tuned models can maintain specific domain-related weaknesses that targeted, resource-efficient continuous pretraining can systematically address. The substantial performance gains achieved with our focused adaptation approach, utilizing significantly smaller datasets than traditional domain specialization methods, demonstrate the viability of efficient domain adaptation strategies that prioritize judicious dataset curation over sheer data volume.

\textbf{Semantic Similarity Analysis.} To assess the contextual understanding capabilities of our domain-adapted models, we conducted semantic similarity evaluation for the 8B and 14B parameter architectures across all three cybersecurity benchmark datasets. During inference, models were prompted to generate explanatory responses of 20-30 tokens justifying their answer selections. We computed cosine similarity scores between the model-generated explanations and the ground-truth correct answers to quantify the degree of semantic alignment.

\begin{figure}[!h]
    \centering
    \includegraphics[width=0.9\linewidth]{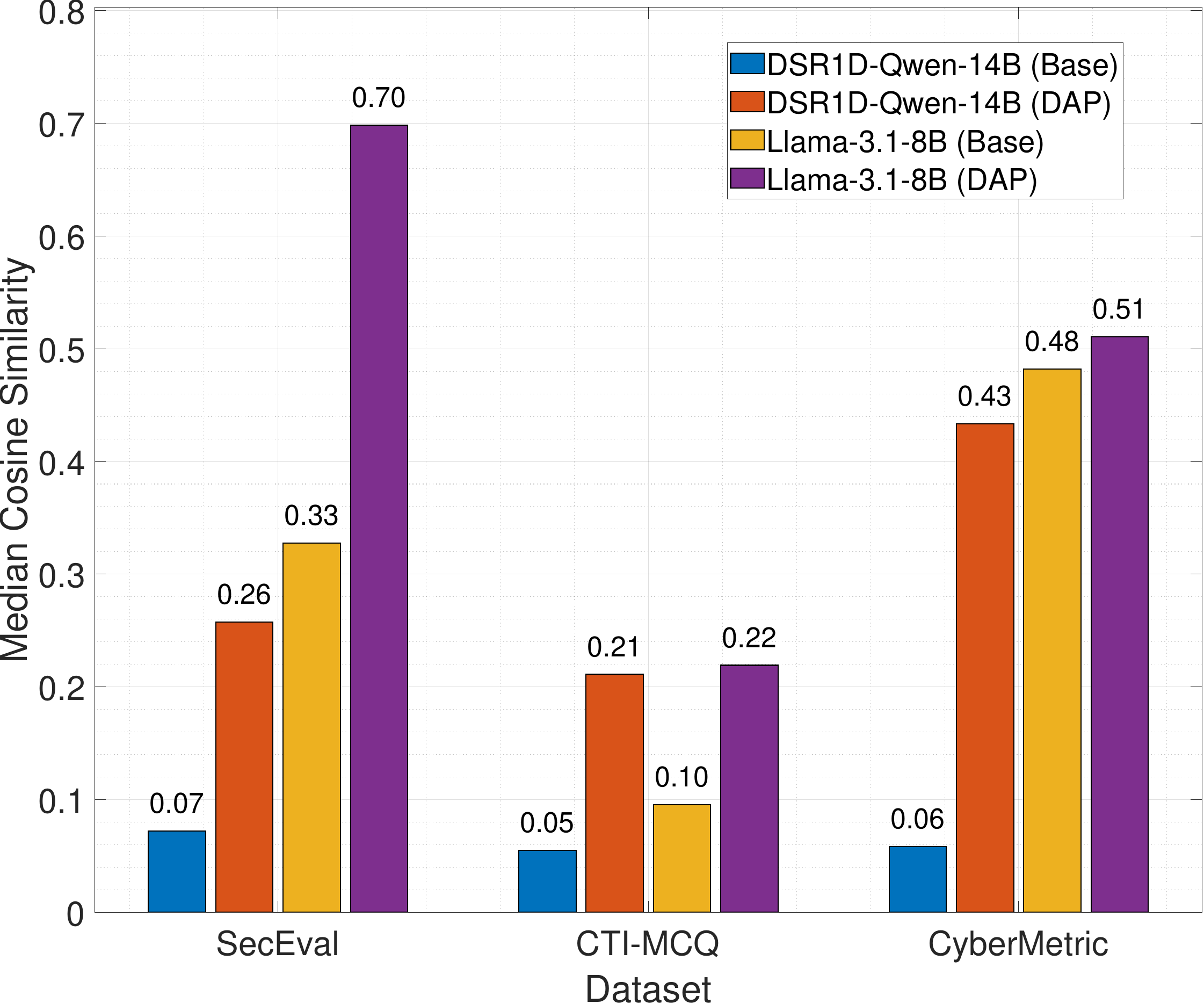}
    \caption{Semantic similarity analysis comparing model explanations to correct answers across cybersecurity benchmarks. Domain-adapted models consistently demonstrate enhanced contextual understanding.}
    \label{fig:semantic_similarity}
\end{figure}

Figure~\ref{fig:semantic_similarity} demonstrates consistent enhancement in contextual understanding following domain adaptation across all evaluated benchmarks. The results confirm that domain-adapted variants consistently achieve higher median cosine similarity scores compared to their baseline counterparts, with the most substantial improvements observed on the SecEval benchmark. Notably, the Llama-3.1-8B-DAP model exhibits an increase from 0.33 to 0.70 on SecEval, representing a 112\% improvement that highlights the profound impact of cybersecurity-focused continuous pretraining on semantic comprehension. It is worth noting that despite having a 0.002 accuracy difference between the base and our \gls{DAP} model (Figure~\ref{fig:acc_cmp_seceval}), the cosine similarity is higher. Hence, indicating that the loss in accuracy might have stemmed from the limited domain knowledge in the small \gls{DAP} dataset rather than the loss in domain understanding.

The DSR1D-Qwen-14B architecture displays more modest but consistent improvements across all benchmarks, with enhancements ranging from 0.05 to 0.06. These findings corroborate our primary performance metrics, providing additional evidence that domain adaptation successfully instills specialized cybersecurity knowledge while enhancing the models' ability to generate contextually appropriate explanations. Resource constraints precluded similar analysis for the 70B model due to the computational overhead associated with extended token generation at scale, as detailed in Section~\ref{sec:dis_limitations}.

These results validate the effectiveness of our domain adaptation approach, demonstrating that targeted fine-tuning on cybersecurity-specific datasets significantly enhances model performance in specialized domains while maintaining computational efficiency compared to training larger, general-purpose models from scratch.

\begin{table*}[!ht]
\centering
\caption{Performance comparison of cybersecurity domain adaptation approaches across CTI-MCQ, CyberMetric, and SecEval. Our \gls{DAP} methodology achieves state-of-the-art results while utilizing substantially smaller training datasets compared to existing specialized frameworks. Models are categorized by foundational training approaches, with "Base Alignment" indicating pre-existing instruction-following capabilities and "Extra Tuning" denoting domain-specific adaptations.}
\label{tab:cyber_benchmark_grouped_column}
\resizebox{\textwidth}{!}{%
\begin{tabular}{|c|l|c|c|c|c|c|c|}
\hline
\textbf{Ref.} & \textbf{Model} & \textbf{Size [B]} & \textbf{CTI‑MCQ} & \textbf{CyberMetric} & \textbf{SecEval} & \textbf{Base Alignment} & \textbf{Extra Tuning} \\
\hline
\multirow{4}{*}{\cite{tihanyi2024cybermetric}}
 & Mixtral‑8$\times$7B‑Instruct      & 45  & --      & 0.911 & --     & SFT + DPO    & None \\*
 & Falcon‑180B‑Chat                 & 180 & --      & 0.871 & --     & SFT (Chat‑style)    & None \\*
 & Mistral‑7B‑Instruct‑v0.2         & 7   & --      & 0.764 & --     & SFT + DPO    & None \\*
 & Gemma‑1.1‑7B‑it                  & 7   & --      & 0.758 & --     & SFT + (novel) RLHF         & None \\
\hline
\multirow{1}{*}{\cite{alam2024ctibench}}
 & Llama‑3‑70B‑Instruct             & 70  & 0.657  & --     & --     & SFT + (multi-round) DPO    & None \\
\hline
\multirow{1}{*}{\cite{alam2024ctibench, tihanyi2024cybermetric}}
 & Llama‑3‑8B‑Instruct              & 8   & 0.613  & 0.731 & --     & SFT + RLHF    & None \\
\hline
\multirow{2}{*}{\cite{li2023SecEvalLeaderboard}}
 & Mistral‑7B‑v0.1                  & 7   & --      & --     & 0.437 &  None (pretrain only)   & None \\*
 & Qwen‑7B                          & 7   & --      & --     & 0.314 & None (pretrain only)    & None \\
\hline\hline
\multirow{1}{*}{\cite{yu2025primus}}
 & Llama‑Primus‑Base (\textcolor{customred}{Cybersecurity specialized})                & 8   & 0.667  & 0.866     & 0.500 & SFT + RLHF  & Continuous Pretraining \\
\multirow{1}{*}{\cite{kassianik2025llama}}
 & Foundation-Sec-8B (\textcolor{customred}{Cybersecurity specialized}) & 8 & 0.676 & 0.851 & -- & SFT + RLHF & Continuous Pretraining \\
\hline\hline
 \multirow{1}{*}{\textbf{Baseline}}
 & Llama‑3‑8B‑Base & 8 & 0.499 & 0.485 & 0.620 & None (pretrain only) & PEFT + Continuous Pretraining \\
\hline\hline
\rowcolor{myLightGray}
 & Llama‑3.1‑8B‑DAP                 & 8   & 0.642  & 0.858 & 0.748 &  None (pretrain only)   & DAP \\*
\rowcolor{myLightGray}
 & DSR1D-Qwen-14B-DAP    & 14  & 0.685  & 0.904 & 0.789 & KD + SFT    & DAP \\*
\rowcolor{myLightGray}
\multirow{-3}{*}{\textit{\textbf{This work}}} & Llama-3.3-70B-Ins‑DAP       & 70  & \textbf{0.718} & \textbf{0.933} & \textbf{0.864} & SFT + RLHF & DAP \\
\hline

\end{tabular}%
}
\end{table*}

%% file: Sections/6_discussion.tex
\section{Discussion}
\label{sec:discussion}
The proliferation of \glspl{LLM} in \gls{NLP} has established their capability for generating contextually coherent responses across diverse domains. However, the computational and data requirements for training domain-specific models from scratch remain prohibitive. We demonstrated that \gls{DAP} offers a viable pathway for specializing general-purpose models to cybersecurity applications in a data-efficient manner, providing foundational capabilities for subsequent task-specific adaptations through \gls{SFT} or \gls{RL}. 

Our investigation employed continuous pretraining rather than adapter-based methods, prioritizing the comprehensive integration of domain knowledge despite the increased computational demands. This approach enables models to extract and comprehend domain-specific semantics from natural language---a prerequisite for complex downstream tasks, including security report analysis and document summarization. The custom dataset, comprising 126 million words from technical documents, standards, and research papers, was deliberately constrained to preserve general language capabilities while instilling cybersecurity-specific nuances.

{\color{customblue}
\textbf{Key Findings.} The empirical evaluation reveals nuanced patterns in model adaptation efficacy across different architectures and dataset scales. Performance improvements on domain-specific benchmarks validate the effectiveness of \gls{DAP}, with the magnitude of gains indicating both successful specialization and opportunities for further optimization through enhanced data diversity and quality.

\begin{itemize}[leftmargin=*]
    \item \textbf{Scale-Dependent Learning Dynamics.} The differential performance across model sizes illuminates critical relationships between architectural scale and training data requirements. The 8-billion-parameter model exhibited performance regression on SecEval (Figure~\ref{fig:acc_cmp_seceval}), which we attribute to the insufficient training data relative to the model's capacity. This model received only 1 million tokens compared to 50-100 times more for larger variants, suggesting that effective domain adaptation requires careful calibration of dataset size to parameter count. This observation underscores the importance of proportional data allocation strategies when adapting models of varying scales.
    
    \item \textbf{Unsupervised Fine-Tuning Constraints.} The moderate performance improvements likely reflect inherent limitations in unsupervised continuous pretraining. Weight updates without explicit supervision may compromise instruction-following capabilities acquired through prior \gls{SFT} and \gls{RL} stages, particularly for complex multi-answer reasoning tasks, such as those in SecEval. Subsequent instruction tuning on curated datasets emphasizing reasoning and task-specific competencies could address these limitations and unlock additional performance gains.

    \item \textbf{Hyperparameter Trade-offs.} Our conservative hyperparameter configuration, particularly the low learning rate ($1 \times 10^{-6}$) and limited training epochs (2-3), was specifically designed to mitigate catastrophic forgetting of general language capabilities. While this approach successfully prevented severe performance degradation on general tasks (as evidenced by the maintenance of instruction-following abilities), it may have constrained the magnitude of domain-specific improvements that could be achieved through more aggressive adaptation strategies. Future work validating general capability preservation through comprehensive benchmarking would enable more informed hyperparameter optimization.

\end{itemize}
}

{\color{customblue}
\textbf{Comparison to Existing Frameworks.} Table~\ref{tab:cyber_benchmark_grouped_column} provides a comprehensive comparative evaluation of our domain adaptation methodology against existing cybersecurity-focused \glspl{LLM} and open-source baseline architectures. All models are evaluated on identical benchmark datasets to ensure fair comparison across different architectural paradigms and training approaches. The comparative analysis reveals several insights into the effectiveness and efficiency of our \gls{DAP} framework.

\textbf{State-of-the-Art Performance.} Our domain-adapted models demonstrate superior performance across all evaluated benchmarks. The \textit{Llama-3.3-70B-Ins-DAP} model achieves state-of-the-art results with accuracies of 0.718 on CTI-MCQ, 0.933 on CyberMetric, and 0.864 on SecEval, surpassing all baseline configurations. \textcolor{customblue}{These results exceed the performance of general-purpose models, including the 45-billion parameter \textit{Mixtral-8x7B-Instruct} and the substantially larger 180-billion parameter \textit{Falcon-180B-Chat}~\cite{tihanyi2024cybermetric}.} This demonstrates that targeted domain adaptation achieves superior effectiveness with more efficient parameter utilization.

\textbf{Data Efficiency Advantages.} The comparison with specialized cybersecurity models reveals particularly notable findings regarding training data requirements. Our approach significantly outperforms \textit{Llama-Primus-Base}~\cite{yu2025primus} across all benchmarks, despite utilizing substantially smaller training datasets: 118.8 million tokens compared to their 2.77 billion-token corpus (a 23-fold reduction). The performance differential is most pronounced on SecEval, where our 70B model achieves 0.864 compared to \textit{Llama-Primus-Base}'s 0.500, representing a 72.8\% relative improvement.

Similar efficiency patterns emerge when compared with \textit{Foundation-Sec-8B}~\cite{kassianik2025llama}, which used approximately 5 billion tokens for training. Despite this substantial data advantage, our \textit{Llama-3.1-8B-DAP} achieves superior accuracy on CyberMetric (0.858 vs. 0.848) while utilizing only 1 million tokens. While \textit{Foundation-Sec-8B} maintains a slight edge on CTI-MCQ (0.662 vs. 0.642), our intermediate \textit{DSR1D-Qwen-14B-DAP} surpasses it (0.685) using just 50 million tokens. This represents a 100-fold reduction in training data compared to \textit{Foundation-Sec-8B}'s 5 billion tokens, demonstrating that strategic dataset curation and model selection can compensate for substantially smaller corpus sizes.

{\color{customblue} \textbf{Domain Adaptation Value.} Comparison with pretrain-only baselines from~\cite{li2023SecEvalLeaderboard} shows the clear value of domain-specific adaptation beyond general pretraining. Our intermediate \textit{DSR1D-Qwen-14B-DAP} not only significantly outperforms established pre-training-only architectures on SecEval (achieving 0.789 compared to 0.437 for \textit{Mistral-7B-v0.1} and 0.314 for \textit{Qwen-7B}), but also delivers improvements over its own base model (0.747). Similarly, \textit{DSR1D-Qwen-14B-DAP} shows notable improvement over its base model. This performance difference indicates that while general language understanding provides a foundation, targeted domain knowledge is essential for complex cybersecurity reasoning tasks.}

\textbf{Intermediate-Scale Efficiency.} The \textit{DSR1D-Qwen-14B-DAP} model exhibits remarkable efficiency, achieving competitive performance across all benchmarks (CTI-MCQ: 0.685, CyberMetric: 0.904, SecEval: 0.789) despite its intermediate parameter size. This effectiveness likely stems from its knowledge distillation foundation, where structured reasoning capabilities inherited from larger teacher models provide enhanced adaptation potential. Compared to instruction-tuned baselines from~\cite{alam2024ctibench}, our 14B model outperforms the \textit{Llama-3-70B-Instruct} on CTI-MCQ while maintaining substantial computational efficiency advantages.

\begin{figure}[!ht]
    \centering
    \includegraphics[width=\columnwidth]{{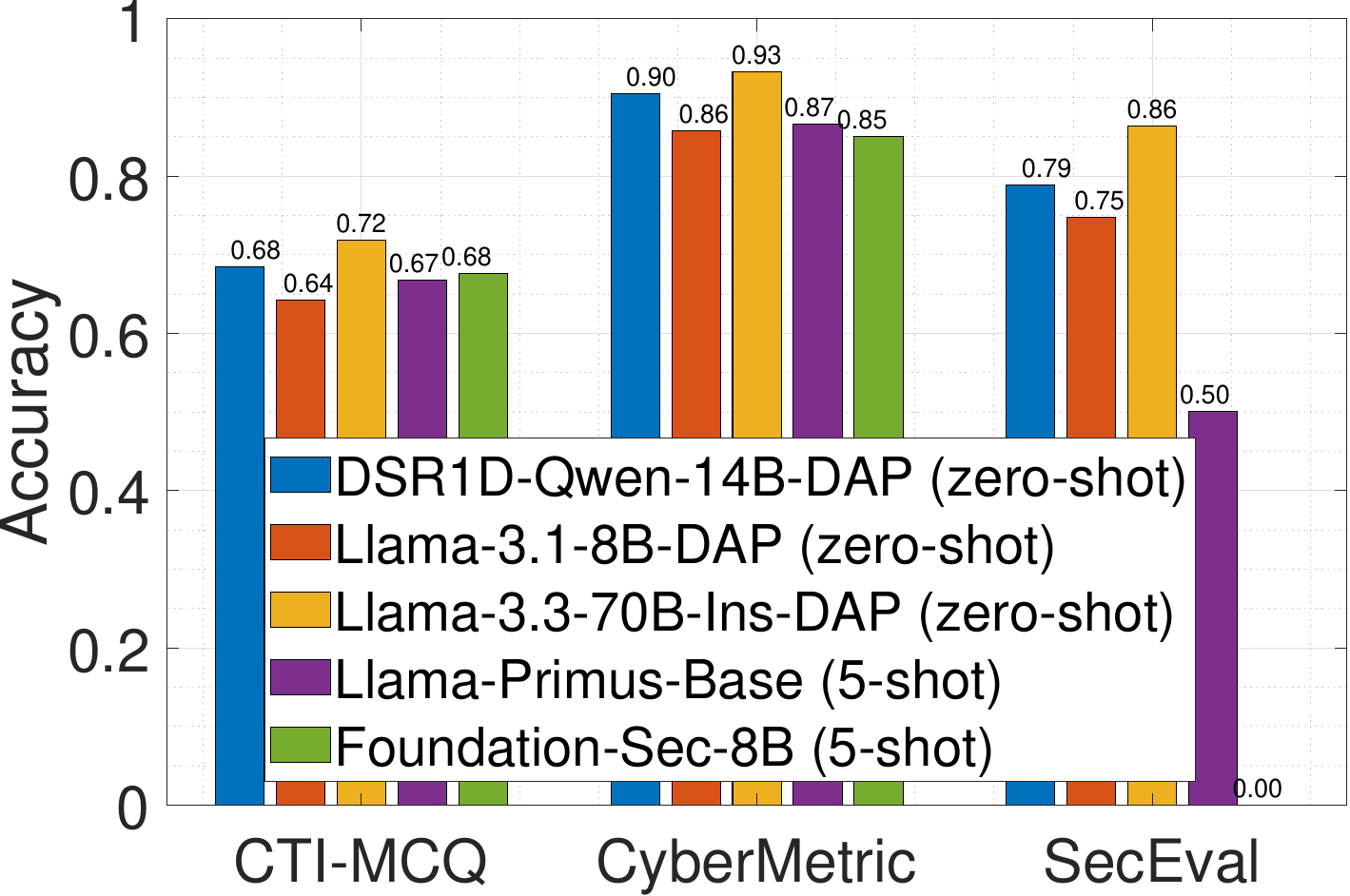}}
    \caption{\color{customblue}{Comparative performance of domain-adapted models against \textit{Llama-Primus-Base} across three cybersecurity benchmarks. Our zero-shot domain-adapted models consistently outperform the specialized 5-shot \textit{Llama-Primus-Base} and \textit{Foundation-Sec-8B}, with \textit{Llama-3.3-70B-Ins-DAP} achieving high performance across all evaluated datasets.}}
    \label{fig:llm_acc_comparison}
\end{figure}

\textbf{Zero-Shot Generalization.} Our models achieve superior performance through zero-shot evaluation, while comparison baselines including \textit{Llama-Primus-Base} and \textit{Foundation-Sec-8B} employ 5-shot prompting strategies. The 5-shot approach provides task format examples that enable the model to infer appropriate response structures augmented through pattern recognition. In contrast, our domain-adapted models achieve competitive or superior results without such assistance. Despite this methodological difference (zero-shot vs. 5-shot evaluation), our \gls{DAP} models demonstrate higher performance across all three benchmarks. This achievement is particularly notable given the substantial disparity in training data volumes, suggesting that careful dataset curation and model selection effectively compensate for limited data availability. Figure~\ref{fig:llm_acc_comparison} visualizes these performance comparisons and highlights the performance of our approach.

{\color{customred}\textbf{Comparison to Parameter-Efficient Methods.} We validated our approach against a \gls{PEFT} baseline using \gls{LoRA}~\cite{hu2021lora}, as presented in Table~\ref{tab:cyber_benchmark_grouped_column}. While adapter-based methods offer computational efficiency, the \gls{LoRA} baseline yielded limited accuracies of 0.499, 0.485, and 0.620 across CTI-MCQ, CyberMetric, and SecEval, respectively. In contrast, our \gls{DAP} approach achieved significantly superior performance (up to 0.642, 0.858, and 0.748). These results confirm that operating on top of frozen base representations is insufficient for deep domain specialization, whereas \gls{DAP} successfully instills foundational cybersecurity awareness. Future research integrating adapter-based methods atop our domain-adapted models could leverage both this comprehensive domain understanding and efficient task-specific adaptation.}}

\subsection{Implications}
\label{sec:dis_implications}

The successful demonstration of efficient domain adaptation carries significant implications for both industrial applications and academic research trajectories.

\textbf{Industrial Use Cases.} The primary application domain centers on enhancing digital system security through generative AI capabilities. When incorporating accepted cybersecurity standards and regulations, adapted models enable automated compliance analysis, identifying policy gaps and conformity issues within corporate frameworks. These \gls{AI} capabilities extend to vulnerability detection and mitigation, offering both syntactic and logical bug fixes to preempt security compromises.

Natural language interfaces enable cybersecurity professionals to perform sophisticated analyses of system logs and configuration files through conversational queries, allowing for seamless interaction. The models' contextual understanding facilitates identification of suspicious activities and vulnerabilities, enabling proactive threat mitigation. Security experts can leverage these capabilities to address potential vulnerabilities before they are exploited, thereby supporting comprehensive security postures against various threats, including identity theft and system compromise.

\textbf{Academic Research.} This paper establishes foundational insights into efficient \gls{LLM} domain adaptation using constrained datasets. The adapted models serve as springboards for future investigations, including the refinement of specialized tasks through parameter-efficient methods, such as adapter-based approaches (e.g., \gls{PEFT} and \gls{LoRA}). Notably, the 14-billion-parameter model achieved performance approaching that of the 70-billion variant, suggesting potential efficiency gains through architectural optimization. This observation warrants systematic investigation across broader evaluation frameworks to validate the generalizability of these findings. The moderate performance differentials may reflect limitations in dataset scope rather than fundamental constraints, indicating opportunities for targeted dataset enhancement strategies.

The demonstrated viability of small-scale domain adaptation challenges prevailing assumptions about data requirements for effective \gls{LLM} specialization, opening new research avenues for resource-constrained model development and deployment.

\subsection{Challenges, Limitations, and Future Directions}
\label{sec:dis_limitations}

\textbf{Computational Resource Constraints.} The primary challenges centered on hardware resource availability for training large-scale language models. The 70 billion-parameter model proved especially demanding, requiring scaling from 8 NVIDIA H100 instances by factors of 2 to 3 to achieve sufficient computational capacity, even in half-precision mode. Such scaling dramatically increased computational costs, while still necessitating a reduction in the context window to manage memory constraints. 

In contrast, models with 8 and 14 billion parameters proved more tractable, though context windows remained constrained to 1024-2048 tokens to avoid memory overflow on lower-tier instances. These reduced context windows inherently limit the models' ability to process extended prompts and extract richer contextual information, which may impact their performance on tasks that require a broader understanding of context. While public cloud platforms offer access to higher-tier GPU resources that could alleviate these constraints, the associated costs substantially exceeded available research budgets.

\begin{figure}
    \centering
    \includegraphics[width=0.8\linewidth]{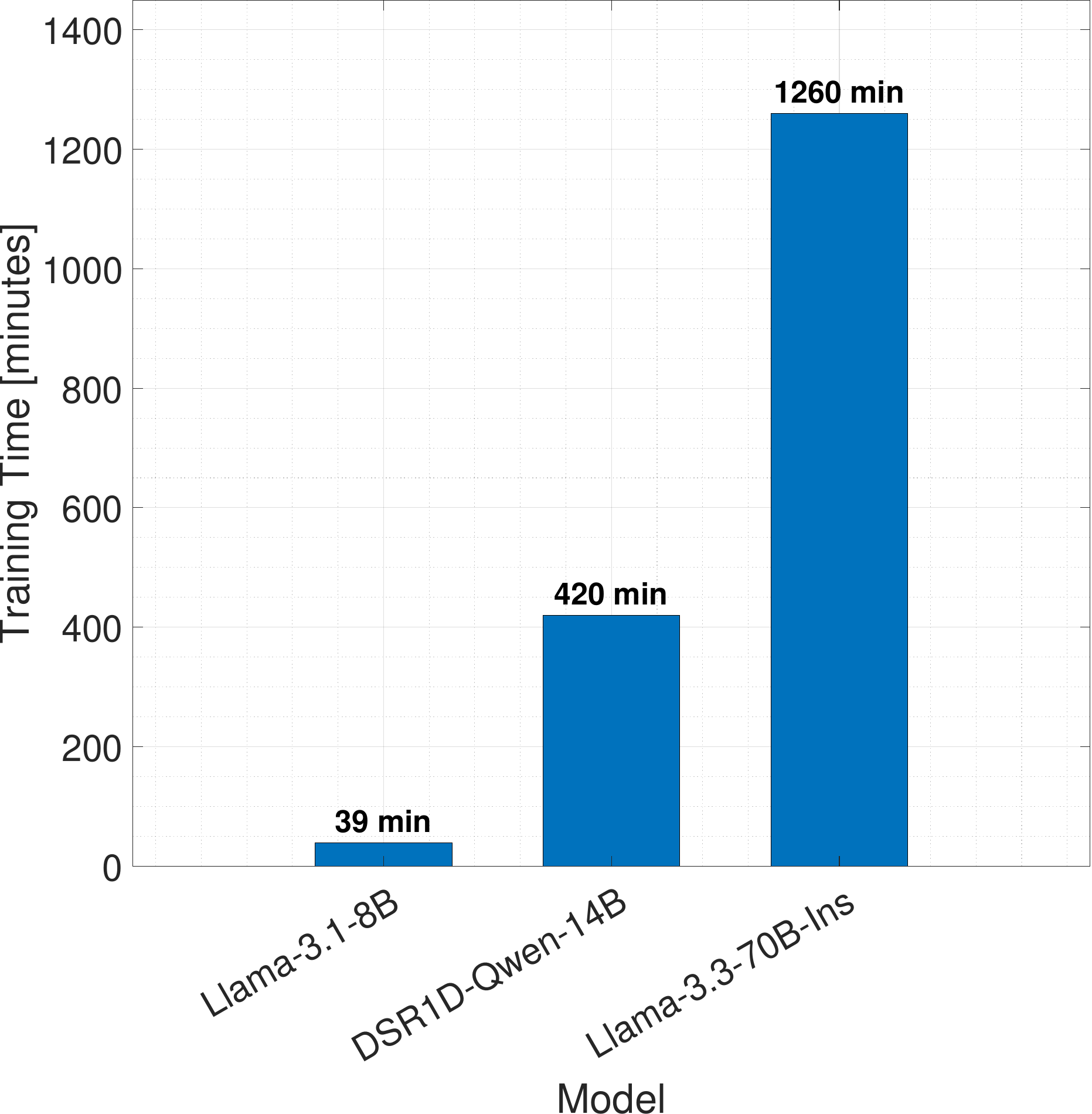}
    \caption{Training time comparison across model architectures demonstrating exponential scaling of computational requirements with model size.}
    \label{fig:dap_time}
\end{figure}

{\color{customblue}
\textbf{Cost-Benefit Analysis.} The instances operate on an hourly billing model where total cost scales linearly with training duration and instance specifications. This cost structure necessitates careful consideration of trade-offs between computational efficiency and resource allocation strategies. Our cost analysis reveals significant scaling dependencies across model architectures. The DSR1D-Qwen-14B model, deployed on dual ml.p5.48xlarge instances (approximately \$63.296 per hour~\cite{awsSageMakerPricing}), required 420 minutes (7 hours) of training time, resulting in domain adaptation costs of \$886.144 (excluding taxes). As illustrated in Figure~\ref{fig:dap_time}, training duration exhibits exponential growth characteristics with respect to model parameters and dataset size, directly impacting computational expenditure. Table~\ref{tab:training_cost_summary} provides a comprehensive breakdown of training costs across all evaluated architectures.
}

The resource scaling relationship becomes particularly pronounced when comparing intermediate and large-scale architectures. While the 14B model demonstrates manageable computational requirements, the 70B architecture necessitates substantially enhanced infrastructure allocation, including extended training durations and higher-tier instance configurations. \textcolor{customblue}{This scaling challenge led to total project costs of approximately \$23,000 (excluding taxes) when accounting for all training experiments and preliminary trials. The final configurations reported in this work, however, cost \$3,581.44 (excluding taxes).}

\begin{table}[!h]
\centering
\caption{\textcolor{customblue}{Total training cost estimation across different model architectures.}}
\label{tab:training_cost_summary}
\resizebox{\columnwidth}{!}{%
\begin{tabular}{|l|c|c|}
\hline
\textbf{Model} & \textbf{Running Time (hours)} & \textbf{Estimated Total Cost (USD)} \\ 
\hline
\textit{Llama-3.1-8B} & 0.65  & \$36.87   \\ 
\hline
\textit{DSR1D-Qwen-14B} & 7.00  & \$886.14  \\ 
\hline
\textit{Llama-3.3-70B-Ins} & 21.00 & \$2,658.43 \\ 
\hline
\end{tabular}%
}
\end{table}

\textbf{Sustainability and Energy Efficiency.} A critical but underexplored dimension of LLM domain adaptation research is its energy and environmental footprint. The computational cost of training large-scale models is proportional to the volume of data processed and the number of gradient steps, both of which scale linearly with dataset size under comparable hyperparameter configurations. Our \gls{DAP} approach, which achieves state-of-the-art cybersecurity specialization with 118.8 million tokens, has a substantially lower energy footprint than existing specialized models. 

To contextualize, Llama-Primus-Base~\cite{yu2025primus} was trained on 2.77 billion tokens---a 23-fold larger corpus---and Foundation-Sec-8B~\cite{kassianik2025llama} on 5 billion tokens, representing a 42-fold increase. Under proportional scaling assumptions, achieving equivalent training coverage with Llama-Primus-Base's corpus on our 70B infrastructure would require approximately 23 times our reported training duration of 21 hours, corresponding to an estimated 483 GPU-hours on 16 NVIDIA H100 instances and a projected cost exceeding \$61,000, compared to our actual expenditure of \$2,658.43. 

While direct energy consumption measurements were not collected in this work due to infrastructure constraints, GPU-hours serve as a well-established metric for energy expenditure in distributed training settings~\cite{Patterson2021}. These projections show that the data-efficiency gains reported in this work are not merely a financial convenience but a tangible contribution to reducing the carbon footprint of \gls{AI} development. We demonstrate that high-quality domain adaptation is achievable with a fraction of the data conventionally assumed necessary; our methodology supports the broader goal of sustainable and responsible \gls{AI}, particularly for organizations operating under real computational and environmental constraints.

\textbf{Limitations.} Several methodological constraints warrant discussion to contextualize our findings appropriately.

\textit{Comparison with Parameter-Efficient Baseline.} \textcolor{customred}{To provide a comprehensive comparative assessment, we evaluated adapter-based baselines using \gls{LoRA}~\cite{hu2021lora} and \gls{PEFT}~\cite{houlsby2019parameter} methods. This baseline evaluation yielded accuracies of 0.499 on CTI-MCQ, 0.485 on CyberMetric-2000, and 0.620 on SecEval (with 790 exact matches). These results fall significantly below the performance achieved by our \gls{DAP} approach, empirically validating our hypothesis in Section~\ref{sec:method_rationale} that \gls{DAP} is essential for instilling intrinsic domain understanding. However, as this evaluation was conducted on an Llama-3.1-8B-Base model, it remains an open question whether larger models could compensate for the limitations of adapter-based methods. Consequently, future work must investigate the impact of different model sizes on \gls{PEFT} performance to determine whether parameter efficiency scales comparably to full-domain adaptive pretraining in cybersecurity contexts.}

\textit{Non-Uniform Dataset Allocation.} \textcolor{customblue}{A further limitation arises from the non-uniform dataset allocation across model scales. The 8B model received 1 million tokens, the 14B model received 50 million tokens, and the 70B model received 118.8 million tokens. This configuration stemmed from both methodological intent and resource feasibility, aiming to examine how varying corpus sizes influence domain adaptation efficiency across different parameter scales. The smaller dataset for the 8B model was deliberately employed to represent a minimal-data scenario for assessing adaptation efficiency boundaries, while larger variants explored scale-sensitive performance characteristics with proportionally increased data availability.}

\textcolor{customblue}{Although this design provided valuable insights into data-performance dynamics, it also introduced variability that constrains direct cross-architecture comparisons. The reduced gains observed for the 8B model on SecEval (Figure~\ref{fig:acc_cmp_seceval}) suggest that the corpus may have been insufficient relative to its parameter capacity, indicating that model scale and dataset size are interdependent factors in achieving effective adaptation. Accordingly, these findings should be interpreted within the context of exploratory efficiency analysis rather than controlled equivalence testing. Future investigations should employ proportionally scaled datasets to enable balanced evaluation across architectures and isolate the independent effects of parameter count on adaptation efficacy.}

\textit{Architectural Heterogeneity.} \textcolor{customblue}{An additional methodological limitation relates to our analysis of scale-dependent learning dynamics. While our investigation evaluated architectures at 8B, 14B, and 70B parameter scales, these models are not from a homologous family. Our evaluation encompasses Llama-3.1-8B (base-pretrained), DSR1D-Qwen-14B (knowledge-distilled), and Llama-3.3-70B-Ins (instruction-tuned), each representing distinct architectural lineages with different pretraining corpora, optimization strategies, and foundational alignments. This architectural heterogeneity introduces significant confounding variables. Consequently, observed performance differences cannot be attributed exclusively to the parameter scale, as they reflect the combined influence of model capacity, pretraining methodology, and base alignment approaches. A rigorous assessment of pure scaling effects would require controlled comparison across a single, consistent model family (e.g., Llama-3.1-8B, Llama-3.1-70B) with identical pretraining and alignment procedures.}

\textit{Catastrophic Forgetting Assessment.} \textcolor{customblue}{A primary limitation of this investigation is the absence of empirical validation for catastrophic forgetting. While our conservative training configuration (specifically the low learning rate of approximately $1 \times 10^{-6}$ and limited training epochs of 2 to 3) was intentionally designed to mitigate degradation of general linguistic capabilities, its success was not quantitatively measured. As noted in Section~\ref{sec:eval_framework}, resource constraints precluded comprehensive benchmarking of our \gls{DAP} variants against their baselines on established general-domain frameworks such as SuperGLUE~\cite{wang2019superglueBenchmark}.}

\textbf{Future Research Directions.} Several promising avenues exist to build upon and extend our foundational results.

\textit{Catastrophic Forgetting Validation.} \textcolor{customblue}{A priority is addressing the primary methodological limitation of this study through rigorous quantitative assessment of catastrophic forgetting. Future work must benchmark our \gls{DAP} variants against their baselines on established general-domain benchmarks (e.g., MMLU~\cite{son2025kmmlu}, SuperGLUE~\cite{wang2019superglueBenchmark}, HellaSwag~\cite{zellers2019hellaswag}). This validation is essential to empirically confirm preservation of foundational language competencies and establish a complete impact profile for our domain adaptation methodology. Such an evaluation would enable more informed optimization of training hyperparameters, potentially allowing for more aggressive adaptation strategies if general capabilities prove to be robustly preserved.}

\textit{Controlled Scaling Analysis.} \textcolor{customblue}{To properly isolate the impact of parameter scale from architectural artifacts, subsequent investigation should be conducted across a homologous model family (e.g., Llama-3.1-8B, Llama-3.1-70B, Llama-3.3-405B) with consistent pretraining and alignment procedures. This controlled comparison would enable definitive conclusions on the optimal model scale for cybersecurity domain adaptation and inform resource-allocation methods for organizations with varying computational budgets.}

\textit{Expanded Model Architectures.} \textcolor{customblue}{Exploring model selection beyond the evaluated Llama-3.1-8B, DeepSeek-R1-Distill-Qwen-14B, and Llama-3.3-70B-Instruct architectures could yield improved outcomes. Particular interest lies in incorporating models with inherent reasoning capabilities such as \gls{CoT} architectures, which may demonstrate enhanced performance on complex multi-step cybersecurity reasoning tasks. Alternative parameter scales across our evaluated points (e.g., 32B, 40B) may reveal optimal efficiency frontiers that balance performance and computational requirements.}

\textit{Dataset Enhancement.} \textcolor{customblue}{Dataset expansion represents a promising avenue for performance improvement. The varying dataset sizes employed in our investigation (potentially insufficient for the 8B model) suggest that broader data collection encompassing wider cybersecurity topics could provide richer contextual and semantic information. Systematic exploration of dataset composition effects (e.g., standards-heavy vs. literature-heavy corpora) could inform optimal curation strategies. Additionally, incorporating more recent cybersecurity content and emerging threat landscapes would enhance the model's relevance to evolving security challenges.}

\textit{Advanced Training Methodologies.} Training methodology refinements, such as knowledge distillation, could optimize the domain adaptation process. Teacher-student architectures that integrate original model knowledge into loss optimization could potentially mitigate concerns about overfitting and catastrophic forgetting. Other promising techniques include curriculum learning (gradually increasing complexity during adaptation) and mixture-of-experts approaches (specialized sub-networks for different cybersecurity subdomains). Resource constraints prevented the exploration of these techniques in the current study; however, they represent valuable directions for enhancing adaptation efficiency and effectiveness.

{\color{customblue} \textbf{Real-World Deployment.} While our benchmark-based evaluation demonstrates foundational cybersecurity knowledge acquisition, validation in operational environments represents a necessary next step. Future work should assess model performance on real-world security workflows, including the analysis of intrusion detection logs, vulnerability assessment reporting, and Security Operations Center (SOC) alert triage. Such an evaluation would establish practical applicability beyond controlled benchmark environments and identify task-specific fine-tuning requirements for production deployment.}

%% file: Sections/7_conclusion.tex
\section{Conclusion}
\label{sec:conclusion}
\color{customblue}We investigated the potential of pretrained \glspl{LLM} in cybersecurity through domain-adaptive training, demonstrating that specialized domain knowledge can be effectively instilled through resource-efficient methodologies. Our approach achieves state-of-the-art performance across established cybersecurity benchmarks, utilizing 118.8 million tokens compared to 2.77-5 billion tokens employed by existing specialized frameworks, representing a 23- to 42-fold reduction in training data requirements. This efficiency gain challenges prevailing assumptions about data volume requirements for effective domain specialization, demonstrating that curation of high-quality materials can compensate for a limited corpus.

\color{customred} While adapter-based \gls{PEFT} methods provided a baseline with limited performance, we used \gls{DAP} to develop foundational domain understanding, essential for contextualizing prompts and generating coherent, logical responses. The experimental results validate our approach across three architectures (Llama-3.1-8B, DeepSeek-R1-Distill-Qwen-14B, and Llama-3.3-70B-Instruct), with the largest model achieving accuracies of 0.718, 0.933, and 0.864 on CTI-MCQ, CyberMetric, and SecEval benchmarks, respectively. These results surpass both the parameter-efficient baselines and specialized models, including Llama-Primus-Base and Foundation-Sec-8B, despite substantially reduced training overhead.

The domain-adapted models developed through this work provide robust foundations for a wide range of downstream applications, including vulnerability and threat analysis, security document summarization, and compliance assessment. Such specialized models offer practical pathways to enhance organizational security postures while maintaining computational feasibility in resource-constrained environments. Future refinement through \gls{SFT} and \gls{RL} techniques could further enhance reasoning capabilities, enabling advanced decision-making through improved semantic understanding and contextual analysis. Our findings establish that efficient, targeted domain adaptation represents a viable alternative to data-intensive pretraining approaches for cybersecurity \gls{LLM} specialization.